\documentclass[conference]{IEEEtran}
\IEEEoverridecommandlockouts
\usepackage{cite}
\usepackage{amsmath,amssymb,amsfonts,amsthm}
\usepackage{algorithmic}
\usepackage{graphicx}
\usepackage{textcomp}
\usepackage{xcolor}
\usepackage{booktabs}
\usepackage{url}
\usepackage{adjustbox}
\usepackage[framemethod=default]{mdframed}
\newtheorem{proposition}{Proposition}
\newtheorem{corollary}{Corollary}
\newcommand{\E}{\mathbb{E}}
\newcommand{\Var}{\operatorname{Var}}
\definecolor{plainbg}{gray}{0.93}
\newmdenv[backgroundcolor=plainbg,linewidth=0pt,innerleftmargin=6pt,innerrightmargin=6pt,
  innertopmargin=3pt,innerbottommargin=4pt,skipabove=4pt,skipbelow=4pt]{plainbox}
\newcommand{\plainlead}{\textbf{In plain words.} }
\begin{document}
\title{Fingerprinting the Dispatchers behind Distributed Model Extraction}
\author{\IEEEauthorblockN{Snigdha Chandan Khilar}
\IEEEauthorblockA{\textit{Independent Researcher}\\
snkhilar@gmail.com}}
\maketitle
\begin{abstract}
Campaigns that distil a frontier model through its API spread their traffic over thousands of accounts, ``load-balanced'' among them. Content classifiers catch many of these accounts, but only those whose prompts give them away. We ask what the rest still reveal through two timestamps per request: when it arrived and when its streamed answer finished. The answer lies in the campaign's \emph{dispatcher}, the client code that decides which key each worker uses next, because each kind leaves its own mark. Workers sharing a round-robin counter, for instance, push a wave of activity around the ring of keys, whose shape we predict with no fitted parameter. Our main use is forensic expansion: from a few accounts that a reliable alarm has named, timing ranks the rest of the campaign. From a 20\% seed it ranked the remaining keys of shuffling, hashing, hedging and company pools at AUC 0.92--1.00 within 20 minutes, among 742 keys of a live two-replica deployment and among 1{,}611--3{,}069 real accounts of a production serving trace, in frozen tests on held-out data. On that trace the frozen detectors left at least 96\% of random account groups unflagged and classified 10 of 11 injected pools. They also found two real relays of 46 and 56 accounts with near-identical request counts (one in a post-hoc rescan), which the untuned classifier named round-robin relays. Blind search for injected rings failed there, and the discoverability law held only with a constant refitted on real noise. An adaptive attacker could not hide a rotating dispatcher even with 30\,s of jitter. Pinning keys escapes, freely under loose quotas but with only 55--61\% of throughput when quotas bind and concurrency is scarce. Of 68 predictions locked in advance, 52 held. Timing shows that accounts share machinery, not why.
\end{abstract}


\section{Introduction}
Distillation through public APIs now happens at industrial scale. One provider reported more than 16 million exchanges carried by about 24{,}000 fraudulent accounts, and in one case a single proxy network managed more than 20{,}000 accounts at the same time~\cite{anthropic2026}. Another provider has described campaigns of more than 100{,}000 prompts~\cite{gtig2026}. The same reports describe how the traffic was organised. One campaign produced ``synchronized traffic across accounts'', in which ``identical patterns, shared payment methods, and coordinated timing suggested `load balancing' to increase throughput, improve reliability, and avoid detection''~\cite{anthropic2026}. Later reports mention a campaign that relayed almost 300{,}000 customer requests through 5{,}380 fraudulent accounts, and another that rotated through 273 accounts~\cite{anthropic2026c}.

Defences have caught up quickly. Classifiers now ban tens of thousands of distillation accounts, ``often in less than two hours'' after an account starts a campaign, with a false-positive rate below 0.1\%~\cite{anthropic2026b}. Providers also mention ``behavioral fingerprinting systems'' and tools ``for identifying coordinated activity across large numbers of accounts'', though they do not say how these work~\cite{anthropic2026}. Two gaps remain. A content alarm names only the accounts whose \emph{questions} gave them away, which is usually a minority of the pool. The calibrated two-scale content test of~\cite{khilar2026} named 12--31\% of a campaign's colluding accounts, at a precision of 0.94--1.00. And a campaign whose prompts look like ordinary users', or which simply relays real users, may never trigger a content alarm at all.

This paper looks at a channel that such campaigns cannot easily switch off: \emph{when} each request on each key starts, and when its answer finishes streaming. Two habits of campaign software shape those timestamps. First, its workers are usually \emph{closed loops}. A worker sends a request, waits for the whole answer, and only then sends the next one. Second, something has to decide which key each request goes out on. We call that piece of code the \emph{dispatcher}. It might be a shared \texttt{itertools.cycle(keys)}, a router such as LiteLLM's, a rule that moves to the next key whenever the provider answers ``429: too many requests'', a fresh key for each agent episode, or a relay that forwards independent users. Users of open-source LLM clients ask for key rotation ``to prevent request limit blocking''~\cite{roocode,opencode}, and ``load balancing'' across accounts is exactly this kind of dispatcher.

\textbf{A small example.} Picture eight workers sharing 40 keys through one round-robin counter. Request 1 goes to key 1, request 2 to key 2, and so on around the ring. At any moment at most eight keys are busy, and they are not scattered at random. They sit next to each other on the ring, like a moving spotlight, and the spotlight keeps sweeping round. Seen from the provider's logs, keys that are neighbours in the client's list tend to be busy together, keys far apart rarely are, and the number of busy keys in the pool stays strangely steady. None of this depends on what the requests say. That regularity is what we measure, and different dispatchers produce different regularities.

\textbf{What we contribute.}
\begin{itemize}
\item \emph{Forensic expansion.} Given a few accounts that some precise alarm has already named (content, payment or metadata), timing ranks the rest of the campaign among all keys within tens of minutes. It picks its own linking statistic without labels (Section~\ref{sec:forensic}).
\item \emph{Signatures, with theory.} We derive four statistics, each one the natural test against the pattern a particular dispatcher creates. For rotation rings we predict the travelling wave, its shape and its harmonics, with no fitted parameter. We also estimate how many workers sit behind a pool (Section~\ref{sec:theory}).
\item \emph{Blind discovery} of rotation rings among all keys, for long-lived campaigns that content never flags. We give a condition, of the BBP type from random-matrix theory, that says how much data a ring needs before it can be found, and an algorithm that never forms the full key-by-key matrix (Section~\ref{sec:discovery}).
\item \emph{An exact active test.} The provider can delay the end of each answer by a secret random amount and watch whether other keys react. This gives exact p-values. We analyse how visible the delay is, how jitter affects it, and what it costs (Section~\ref{sec:wm}).
\item \emph{Limits, with costs.} Pinned keys and steady open loops are provably invisible to timing. A content--timing trade-off shows what pinning costs against a per-key content monitor, and when that cost disappears. We measure the price of every other evasion (Section~\ref{sec:limits}), including for an adaptive attacker on production traffic (Section~\ref{sec:adaptive}).
\item \emph{Evidence under locked predictions.} Ten experiments: simulation, 1.96 million WildChat requests, five live vLLM runs, and 2.2 million requests from three hours of the Chutes production trace, two of them held out with frozen code. The live runs include server stress, two replicas, per-key quotas, and dispatchers we had never seen when the thresholds were set. The production hours contain real rotation pools, which the frozen classifier names. We report every prediction that failed (Appendix~\ref{app:ledger}).
\end{itemize}
Code, request logs and locked predictions are available at \url{https://github.com/nssprogrammer/dispatcher_fingerprinting}.

\textbf{How to read this paper.} Section~\ref{sec:primer} explains every idea we use in everyday terms, before the formal results. After each formal result, a shaded box marked ``In plain words'' says what the result means and why it is true. A reader who skips the mathematics can follow the paper through those boxes and the tables.

\section{Related work}
\textbf{Defences against extraction.} Extraction through prediction APIs~\cite{tramer2016} and distillation from generated data~\cite{hinton2015} led to per-account monitors such as PRADA~\cite{juuti2019}, and to traffic-window tests for LLM APIs~\cite{liu2026}. Herley shows how to bound the abusive share of a traffic mixture without labels~\cite{herley2026}. Khilar~\cite{khilar2026} measures how far content-only monitoring can go and introduces randomised throttles for exact failover tests~\cite{khilar2026}. Providers deploy content classifiers, behavioural fingerprints and coordination detectors~\cite{anthropic2026,anthropic2026b}, but their methods are not public. We offer an open timing method, with its theory and its limits, and we place it behind those alarms rather than in competition with them.

\textbf{Coordination and timing.} Lockstep and synchrony detectors look for accounts that act at the same moments~\cite{beutel2013,cao2014}. BotMiner cross-correlates communication and activity clusters~\cite{gu2008}, and studies of sock puppets look at posting times~\cite{kumar2017}. Our signals point the other way. Keys driven by a fixed set of workers \emph{take turns} rather than acting together. Stepping-stone detection correlates the ends of idle periods across network connections~\cite{zhang2000}. Active flow watermarks embed a pattern of delays and look for it downstream~\cite{wang2003,yu2007,houmansadr2009}, where staying invisible and surviving tampering are the main concerns. Our echo appears on \emph{other accounts} at the same server, and we study both concerns (Section~\ref{sec:wmadv}).

\textbf{Timing side channels in LLMs.} Token lengths and timing let an eavesdropper infer what encrypted LLM traffic says~\cite{weiss2024,carlini2024}, so timing and content are not independent. The provider already sees the content, so for it these channels add nothing. For our detectors they mean that answer durations carry some content, whereas the dispatch pattern we test is set by the client.

\textbf{Statistics.} In ecology, the same variance ratio detects species that compensate for each other~\cite{schluter1984,hallett2014}. Our guiding analogy is antibunched light, in which photons arrive more evenly than chance~\cite{kimble1977,hbt1956}. Counting hidden units from fluctuations goes back to ion-channel noise analysis~\cite{sigworth1980}, and our worker count is an operational law of queueing systems~\cite{denning1978}. Closed and open workloads behave differently~\cite{schroeder2006}. Recovering ring order is circular seriation~\cite{recanati2018}. Blind discovery rests on the BBP transition~\cite{baik2005,benaych2011}, and the watermark on Fisher's randomisation test~\cite{fisher1935}.

\section{Setting and threat model}\label{sec:setting}
For each request $r$ the provider logs its key $a_r$, its arrival time $s_r$, the moment $c_r$ at which the model finished generating, and the moment $e_r$ at which the last byte left the server. Without a watermark, $e_r=c_r$. A \emph{pool} is any set of $M$ keys that we examine together. The \emph{occupancy} of key $i$ at time $t$, written $X_i(t)$, is the number of its requests in flight, that is, requests with $s_r\le t<e_r$, and $N=\sum_iX_i$ is the total for the pool. A closed-loop worker sends a request, waits for the stream to end, pauses for a client gap $G$, and sends again. $D$ denotes the duration of an answer and $\lambda$ the rate at which the pool dispatches requests. The provider sees keys, never workers. Table~\ref{tab:pops} lists the dispatchers we test.

\textbf{The attacker} controls $K$ concurrent workers and $M$ keys. It may relay real users, choose any dispatcher, add random waits (jitter), hold keys for many requests, or pin each worker to a key. It does not control the server, but it sees everything its own clients receive, including any delay we add.

\textbf{The defender} is the provider. It sees timestamps, keys and, if it wishes, content. It may delay the end of a stream by a secret random amount. We assume no payment, network or device metadata; when such metadata exists it only helps the defender.

\textbf{Content regimes.} When we combine timing with content (Table~\ref{tab:two}), we follow~\cite{khilar2026} and consider four kinds of campaign prompt. \emph{Dataset} campaigns send problems from a public maths dataset. \emph{Half-padding} campaigns mix those problems half and half with real user prompts. \emph{Tilted} campaigns send real user prompts, re-weighted towards the useful topic; this is the least detectable campaign of a given size. \emph{Written} campaigns send problems that a model has rewritten in users' voice. Their value to the attacker is measured as the GSM8K accuracy that a 360M-parameter student gains after fine-tuning on the campaign's 3{,}000 question--answer pairs: $+10.4$, $+7.6$, $+2.2$ and $+4.4$ points respectively. Content detection is the weekly rate at which hollowed spectral selection over sentence embeddings flags the campaign, at about 2\% false alarms. These numbers come from~\cite{khilar2026} and appear only in Table~\ref{tab:two}; everything else in this paper is measured here.

\begin{table}[t]\centering\small
\caption{Populations. $K$ workers over $M$ keys. P: Baseline, Repeat and Stress runs; E: Evasion run; F: Two-replica run.}\label{tab:pops}
\setlength{\tabcolsep}{4pt}
\begin{adjustbox}{max width=\columnwidth}
\begin{tabular}{@{}lllcc@{}}
\toprule
Pool & Client & Dispatcher & $K$ & $M$\\
\midrule
humans & asyncio & one key each & -- & 400\\
batch & OpenAI SDK & one key each (benign) & 4 & 4\\
P1 & LiteLLM Router & \texttt{simple-shuffle} & 6 & 40\\
P2 & LiteLLM Router & \texttt{least-busy} & 10 & 40\\
P3 & OpenAI SDK & shared \texttt{itertools.cycle} & 12 & 40\\
P4 & OpenAI SDK & one cycle per worker & 8 & 40\\
P5 & OpenAI SDK & shared cycle + 0--6\,s jitter & 8 & 40\\
P6 & agents & key per episode, tool calls & 6 & 32\\
P7 & relay & open loop, random key & -- & 40\\
P8 & relay & open loop, round-robin & -- & 40\\
C & LiteLLM Router & legitimate company, shuffle & 4 & 10\\
P9 & OpenAI SDK & pinned keys & 8 & 8\\
\midrule
E0--E3 & OpenAI SDK & shared cycle, jitter 0/2/6/15\,s & 8 & 20\\
E4 & OpenAI SDK & shared cycle, 15\,s jitter & 24 & 20\\
E5, E6 & OpenAI SDK & key held for 5 / 20 requests & 8 & 20\\
E7, E8 & OpenAI SDK & random key & 8 / 18 & 20\\
E9 & OpenAI SDK & pinned keys & 8 & 8\\
E10, E11 & relay & open loop, steady / bursty & -- & 20\\
\midrule
F1 & OpenAI SDK & rotate on 429, shared pointer & 8 & 40\\
F2 & OpenAI SDK & rotate on 429, per-worker pointer & 8 & 40\\
F3 & OpenAI SDK & consistent hash of the task ID & 8 & 40\\
F4 & OpenAI SDK & sticky key, 60\,s time limit & 8 & 40\\
F5 & OpenAI SDK & hedged: 2 keys, first answer wins & 6 & 40\\
F6--F9, C & & references: P3, P1, P9, E10, C & & \\
\bottomrule
\end{tabular}
\end{adjustbox}
\end{table}

\section{The ideas in plain words}\label{sec:primer}
This section explains, without formulas, every idea the rest of the paper relies on. Readers who know the statistics can skip it; the formal statements follow in Section~\ref{sec:theory}.

\textbf{Keys, pools and occupancy.} A key is an account's API credential. Draw a timeline for each key and shade it whenever the key has a request in flight. That shaded timeline is the key's \emph{occupancy}. A pool is just a group of keys we look at together: forty keys picked at random, forty keys a classifier flagged, or forty keys we suspect belong to one campaign.

\textbf{Closed and open loops.} A \emph{closed-loop} worker behaves like a person at a counter who asks one question, waits for the full answer, and then asks the next. Most scripts work this way, because they need each answer before they can store it. An \emph{open-loop} sender fires requests on a schedule without waiting, the way a relay forwards whatever its own users send. The difference matters because a closed-loop worker's next request is tied to the moment its previous answer ended, and that tie is visible in the logs.

\textbf{Dispatchers.} The dispatcher is the rule that picks a key for each request. The common ones are:
\begin{itemize}
\item a \emph{shared counter}, in which all workers take the next key from one round-robin list (key 1, 2, 3, \dots, 40, 1, 2, \dots);
\item \emph{per-worker counters}, in which each worker walks the list on its own;
\item a \emph{shuffle}, which picks a key at random each time;
\item \emph{least-busy} routing, which picks the key with the fewest requests in flight;
\item \emph{sticky} routing, which keeps a key for a while, or for a number of requests, and then moves on;
\item \emph{rotate on 429}, which keeps using a key until the provider refuses it for exceeding its quota, then moves to the next;
\item \emph{consistent hashing}, which maps each task to a key by hashing its ID;
\item \emph{hedging}, which sends each request on two keys and keeps the first answer;
\item \emph{pinning}, in which each worker owns one key and never touches another;
\item a \emph{relay}, which forwards independent users' requests on keys chosen at random or in turn.
\end{itemize}
A legitimate company that spreads one job queue over several of its own keys uses a dispatcher too. We include one (pool C) because it looks exactly like a campaign.

\textbf{Turn-taking and the occupancy ratio $R$.} Suppose eight workers share forty keys. Each key flickers on and off, but the pool as a whole never has more than eight requests in flight, and the number stays remarkably steady, because whenever one key goes idle the worker that left it soon starts a request elsewhere. The keys \emph{take turns}. Compare forty independent users: when each one is busy has nothing to do with the others, so the total wanders up and down as much as the individual flickering would suggest. The occupancy ratio $R$ makes this exact. It divides how much the pool's total occupancy varies by the sum of how much each key's occupancy varies. Independent keys give $R=1$. Keys that take turns give $R<1$, because their ups and downs cancel. Keys that are busy at the same moments give $R>1$, which we call \emph{bunching}. Think of eight waiters serving forty tables: individual tables come and go, but the number of tables being served at once barely changes. The ratio can also be computed after removing slow trends: we cut time into blocks of length $\tau$ (say 20 seconds) and subtract each block's own average, so that only faster ups and downs remain. This is what we mean by \emph{high-pass}, and the result is $R_\tau$. A dispatcher that holds each key for a minute looks independent when $\tau$ is a few seconds and turn-taking when $\tau$ is several minutes, so $R_\tau$ also tells us how long keys are held.

\textbf{Correlation.} The correlation between two keys measures whether they tend to be busy at the same moments ($+$), at different moments ($-$), or neither (about 0). With a shared counter, keys next to each other in the client's list are busy at overlapping times, because the counter visits them one after the other while earlier answers are still streaming. Keys far apart on the list are rarely busy together. Plotting correlation against distance on the ring gives a bump around distance zero. We call it the \emph{ring profile}, $c(d)$, and Section~\ref{sec:theory} predicts its shape from the answer durations and the delays between visits.

\textbf{The travelling wave, eigenvalues and Fourier modes.} Write all the pairwise correlations of a pool into a table with one row and one column per key. For a shared counter, if the keys were listed in ring order, every row of the table would be the previous row shifted along by one place, because the counter treats every key alike. Tables with this property are called \emph{circulant}. Their natural building blocks are waves that go round the ring once, twice, three times and so on, which are called Fourier modes. The table's \emph{eigenvalues} say how strongly each wave is present. For a ring, the wave that goes round once is usually the strongest, and its strength is called $\theta_1$. It comes as a pair, a sine and a cosine, and together they give each key an angle, which is its position on the ring. This is how we recover the order of the client's key list without ever being told it. When the bump $c(d)$ is narrow, faster waves (twice round, three times round) are also strong. These are \emph{harmonics}, and they explain some extra eigenvalues we saw in our data.

\textbf{Successors and hand-offs.} When an answer ends on key $i$ and, a moment later, a request starts on key $j$, we call that a \emph{hand-off} from $i$ to $j$. Under per-worker counters, the worker that just finished on key $i$ always moves to key $i+1$, so key $i$ has one favourite successor. \emph{Successor concentration} measures how strongly a key's hand-offs favour a single partner. A shared counter does not have this property, because the next key depends on what all the other workers did in the meantime.

\textbf{Counting workers.} If a pool dispatches two requests per second and each worker's cycle takes four seconds (a three-second answer and a one-second pause), then eight workers must be running, because each worker can only manage one request every four seconds. This is Little's law from queueing theory, and it assumes the workers are always busy, which is not true of agents that pause for tool calls. The only hard part is measuring the pause, which we read from the extra hand-offs that follow each answer.

\textbf{Is it chance? $z$-scores and references.} Every statistic needs a reference for how big it gets by chance, and we use three kinds. For the ring and successor statistics we slide each key's timeline by its own random amount; each key keeps its own rhythm, but any link between keys is destroyed, and we repeat this about forty times. For the occupancy ratio we compare against random pools of ordinary users (the \emph{human null}). For hand-off counts we compare against what a constant rate of starts would give. The $z$-score says how many standard deviations the real value lies from its reference. By convention $z=3$ is unusual; we alarm at $z>6$, which the references almost never reached in our synthetic and light traffic, though heavy production traffic needs its own threshold (Section~\ref{sec:chutes}). We report $z_R$ so that turn-taking is positive, which makes bunching show up as large \emph{negative} values. A test is \emph{anti-conservative} when its reference underestimates chance and so fires too often.

\textbf{Each statistic is the right test for its dispatcher.} For each dispatcher we ask which test would be most sensitive to a small amount of exactly the pattern that dispatcher creates. Statisticians call such a test a \emph{score test}. Like a radio tuned to one station, it is most sensitive to that one pattern. Our ring and successor statistics are tractable approximations to these tests. Random dispatch leads to the occupancy ratio, a shared counter to the ring statistic, and per-worker counters to successor concentration. So the statistics are not arbitrary choices; each is matched to its dispatcher.

\textbf{Hollowing.} When a server gets busy, every answer slows down at once, which makes every key look slightly correlated with every other. \emph{Hollowing} removes that site-wide common movement, together with each key's correlation with itself, so that only the structure between particular keys remains.

\textbf{Finding a ring among thousands of keys.} With no suspects at all, we compute the correlation table for every key at the site and look for the ring's wave in it. The difficulty is noise. Even unrelated keys show small chance correlations, and among $N$ keys these combine into a background whose largest eigenvalue grows like $\sqrt N$. Watching for longer helps, because chance correlations shrink as the window $T$ grows. A ring becomes visible once its strength $\theta_1$ beats the background, and that sets a minimum observation time, $T^*$, which grows in proportion to the number of keys and falls with the square of the ring's strength. This threshold effect is a known result of random-matrix theory, called the BBP transition; our version is calibrated with a measured constant, $\kappa$. On a plot of the eigenvalues (Fig.~\ref{fig:spec}) the noise forms a cloud that ends at the \emph{bulk edge}, and a ring is a pair of eigenvalues sticking out above it. The pair has two nearly equal members, the sine and the cosine wave, which is why we call it \emph{degenerate}. A recovered group is \emph{pure} when it contains no keys from outside the ring. Think of picking out a tune in a crowded room: more people make more noise, and listening for longer lets the tune emerge.

\textbf{The hand-off graph.} A second way to search the whole site is to count, for every pair of keys, how often an answer on one is followed within a second by a request on the other, and to link pairs that do so far more often than chance would allow. Linked keys form groups.

\textbf{The watermark.} The provider can run an experiment. For each request it flips a secret coin, and on heads it holds back the end of the answer by a fraction of a second. If the same worker sends its next request on a \emph{different} key, that next request is delayed as well, and only after heads. Comparing what happens after heads and after tails links the two keys. Because the provider flipped the coins itself, it can compute an \emph{exact} p-value by reassigning the heads and tails many times on its own computer and asking how often chance alone produces an effect as large. This is Fisher's randomisation test. The smallest p-value $B$ reassignments can give is $1/(B+1)$, the \emph{permutation floor}.

\textbf{p-values, error rates and power.} A p-value is the probability of seeing an effect at least as large as ours if nothing were going on. If we accuse whenever $p<\alpha$, then honest users are wrongly accused at most a fraction $\alpha$ of the time (the \emph{level}), and the \emph{power} is the fraction of real campaigns we catch. When we run many tests, some will look significant by luck. The \emph{Holm} procedure keeps the chance of even one false accusation below $\alpha$, and \emph{Benjamini--Hochberg} keeps the expected share of false accusations among all accusations below a target $q$. Bonferroni is the simplest version of the first: divide $\alpha$ by the number of tests.

\textbf{AUC and precision.} When expansion ranks keys, the AUC is the probability that a randomly chosen campaign key is ranked above a randomly chosen unrelated key. An AUC of 0.5 is a coin toss and 1 is a perfect ranking. \emph{Precision at pool size} answers a more practical question: if the campaign has 40 keys and we take the top 40 of the ranking, what fraction really belong to it? Precision depends on the \emph{base rate}, how common the thing we look for is among everything we search.

\textbf{Why some campaigns are invisible.} If each worker is pinned to its own key, the logs show a set of keys each used by one steady client. That is exactly what honest batch jobs look like. No timing test can tell the two apart, because the data look the same in every respect, not because our tests are weak. The same is true for an open loop that sends at random moments on random keys.

\textbf{Fano counts.} The \emph{Fano factor} is the variance of a count divided by its mean. Applied to the number of requests in flight, it gives a second, quick count of workers: $K$ workers each busy part of the time keep the total less variable than a random count, and the mean divided by one minus the Fano factor recovers $K$.

\textbf{Equal request counts.} A round-robin counter gives every key in its list the same number of requests, give or take one. Random choice spreads the counts out much more widely. The $\chi^2$ dispersion test asks whether a group's counts are \emph{too even} to be chance, which gives independent evidence of a counter.

\textbf{Locked, frozen, held out, post hoc.} Before each experiment we wrote down our predictions; for the production trace we also recorded a cryptographic hash of them, so that any later change would be detectable. \emph{Frozen} code is code fixed before the run or the data that tests it. Each detector carries a version number (v0.4 to v0.9), and each version was frozen before the run that tested it; predictions are labelled M1--M9 in the Two-replica run and R1--R9 in the Production study. A \emph{held-out} window is data we did not look at while building anything. A \emph{post-hoc} analysis is one we decided to run after seeing results. We report post-hoc analyses, but always label them as such.

\section{Signatures and theory}\label{sec:theory}
\textbf{Turn-taking.} The occupancy ratio is $R=\Var N/\sum_i\Var X_i$. It is 1 for independent keys and for open-loop Poisson arrivals on random keys, but $(1-p)/(1-p/M)$ for $K$ workers that are each busy a fraction $p$ of the time (the duty cycle) and are sent to random keys (Appendix~\ref{app:proofs}). We also compute $R_\tau$ on high-pass blocks of length $\tau$. A key held for stints of length $L$ looks independent when $\tau\ll L$ and turn-taking when $\tau\gg L$.
\begin{plainbox}\plainlead
With 8 workers each busy 80\% of the time over 40 keys, $R=(1-0.8)/(1-0.02)\approx0.2$: the variance of the pool's total is only a fifth of what independent keys would give. The busier the workers, the more the total is pinned in place and the smaller $R$ gets. An open loop does not wait for answers, so it creates no compensation, and $R$ stays at 1.
\end{plainbox}

\textbf{The travelling wave.} A shared counter sends request $n$ to key $n\bmod M$. Key $i$ is visited once per ring period $T_c=M/\lambda$, and key $i+d$ a random time $S_d$ later.
\begin{proposition}[Ring profile]\label{prop:ring}
If $K<M$ (duty $p=\E D/T_c$) and durations are independent, keys $d$ apart on the ring correlate at
$c(d)=\big(\E\,O(S_d;D,D')/T_c-p^2\big)/\big(p(1-p)\big)$,
where $O(\tau;D,D')=(\min(D,\tau+D')-\max(0,\tau))_+$ is the overlap of two on-intervals that start $\tau$ apart. The hollowed pool correlation matrix is approximately circulant, with eigenvalues $\theta_k=\sum_{d=1}^{M-1}c(d)\cos(2\pi kd/M)$. The top pair, $k=\pm1$, places each key at its ring position as an angle.
\end{proposition}
\begin{plainbox}\plainlead
Two keys are busy together exactly when their answers overlap in time. Key $i+d$ is started a little after key $i$, so whether their answers overlap depends only on that delay and on how long the answers last. The formula averages this overlap and turns it into a correlation. Nearby keys overlap often, distant ones almost never, which gives the bump-shaped profile. Because the counter treats all keys alike, the correlation table is approximately circulant (exactly so in the limit of a long window), its eigenvalues are the Fourier sums $\theta_k$, and the two leading eigenvectors work like a compass that reads off each key's place on the ring. Every input is measured from the pool itself, so nothing is fitted.
\end{plainbox}
\begin{corollary}[Harmonics]\label{cor:harm}
The profile $c(d)$ has an effective width of about $\lambda\E D$ keys. Narrow profiles, as produced by heavy jitter or slow rings, have flat spectra, so the higher harmonics $k=\pm2,\dots$ can also clear a detection threshold.
\end{corollary}
\begin{plainbox}\plainlead
$\lambda\E D$ is how many requests the counter sends out while one answer is still streaming. That is how many neighbours a key overlaps with, and so the width of the bump. A sharp, narrow bump contains many fast waves, just as a short click contains many high notes, so a jittered ring shows up as several eigenvalue pairs rather than one.
\end{plainbox}
Open-loop arrivals change only the distribution of $S_d$, so a round-robin relay shows a ring even though its $R\approx1$. The proposition needs $K<M$, so that a key holds at most one request at a time. With $K\ge M$ a key can hold two, and we have no closed form. We measured that case instead: with 24 workers over 20 keys (E4) the ring was still found at $z=29.9$, while $R$ lost its turn-taking. Shared batching makes durations depend on load, which breaks the independence assumption. It adds a common mode, and hollowing removes it (Section~\ref{sec:crosstalk}).

\begin{proposition}[Successors]\label{prop:succ}
After an answer on key $i$, the worker's next request goes to key $i+\Delta$, with $\E\Delta=K$ for a shared counter and $\Delta=1$ for per-worker counters.
\end{proposition}
\begin{plainbox}\plainlead
With a shared counter, the other $K-1$ workers each take one key while a worker is busy, so its next key lies about $K$ places further on. With per-worker counters, every worker simply moves to the next key in its own list. So the typical jump along the ring tells us the number of workers, and a jump of exactly one points to per-worker counters.
\end{plainbox}
\emph{Successor concentration} is the share of a key's excess end$\to$start hand-offs that go to its most frequent successor, compared with key-wise time-shifted surrogates. The \emph{collision ratio} is the time-integrated number of concurrent pairs on the same key, relative to its value if concurrent requests were spread uniformly over the keys.

\begin{proposition}[Worker count]\label{prop:K}
For $K$ always-busy closed-loop workers, $K=\lambda(\E D+\E G)$~\cite{denning1978}. $\E G$ is read from the excess of end$\to$start lags over a uniform-rate expectation.
\end{proposition}
\begin{plainbox}\plainlead
This is the ``two requests a second, four seconds per cycle, so eight workers'' argument of Section~\ref{sec:primer}. Shortly after an answer ends, more requests start than the pool's average rate would explain; those extra starts are the worker coming back after its pause, and their timing gives the length of the pause.
\end{plainbox}
Appendix~\ref{app:proofs} gives batch-means intervals. Sampling error is small: $\pm0.03$--$0.8$ workers over 36--56 minutes, or $\pm1.3$ with 6\,s of jitter. The larger error is a positive bias of up to 0.4 workers, because we read $\E G$ at the midpoints of lag bins.

\textbf{Every statistic is a score test.} Approximate a pool's high-pass occupancy by a Gaussian process with correlation $I+\epsilon A$, where $A$ describes the dispatcher. The score test for $\epsilon=0$ uses $\mathrm{tr}(A\hat C)$. Random dispatch gives the sum of off-diagonal correlations, which is a weighted form of $R-1$. A shared counter gives a circulant $A$ with an unknown key order, and the spectral relaxation of that problem is the ring statistic. Per-worker counters give a matching on the hand-off matrix, and its row-by-row relaxation is successor concentration (Appendix~\ref{app:score}).
\begin{plainbox}\plainlead
$\mathrm{tr}(A\hat C)$ adds up the observed correlations, weighted by where the dispatcher says correlation should be. It works like a template: lay the expected pattern over the data and see how well it fits. We do not know the client's key order, so we search over orders, and the eigenvector computation is the efficient way to do that search approximately.
\end{plainbox}

\textbf{Classifier.} We fixed all thresholds in simulation before any live run. The classifier is a short decision list:
\begin{enumerate}
\item If ring $z>6$, the pool is a shared rotation. It is a closed loop if $R$ or $R_{20}$ lies more than 3 sd below the human null, and an open-loop relay otherwise.
\item Otherwise, if successor concentration $z>6$, it is a per-worker rotation.
\item Otherwise, if the keys take turns, it is sticky when $1-R_5<\tfrac12(1-R)$, least-busy when the collision ratio is below 0.5, and a random shuffle otherwise.
\item Otherwise, it is bunching if $R$ lies more than 3 sd above the null, and none if not.
\end{enumerate}
Version~0.6 computes the ring statistic at 20, 60 and 180\,s blocks and takes the maximum. Version~0.7 confirms a bunching call on request \emph{start} counts (Section~\ref{sec:crosstalk}). Each version was fixed before the live run that tested it.

\section{Forensic expansion}\label{sec:forensic}
A provider that bans accounts on a content alarm already holds a precise seed. The alarm's precision was 0.94--1.00 in~\cite{khilar2026}, and the deployed classifier's false-positive rate is below 0.1\%~\cite{anthropic2026b}. The question is which other keys belong to the same campaign. Given a seed set $\mathcal{S}$ of keys, we score every other active key $j$ in three ways. The \emph{turn} score is minus the projection of $j$'s standardised high-pass occupancy onto the seed's summed occupancy; random dispatch makes the keys of a pool anti-correlated, so a key that is busy when the seed is idle scores high. The \emph{hand-off} score is the excess of end$\to$start hand-offs within 1\,s between $j$ and the seed, in both directions, over its uniform-rate expectation, expressed as a $z$-score. The \emph{co} score is the projection itself, for pools that bunch.

The choice among the three is made \emph{without labels}. We hold out each seed member in turn, rank it against all non-seed keys under each score, and use the score under which the held-out members rank highest. Expansion needs only one pass over the seed's neighbourhood, with no site-wide eigendecomposition, and because it only starts once an alarm has fired, the base rate of benign shared dispatch (Section~\ref{sec:deploy}) does not dilute its precision the way it dilutes blind discovery.
\begin{plainbox}\plainlead
Once we know a few keys of a campaign, we ask which other keys fit around them: busy when the seed is idle (they take turns with it), starting just after the seed's answers end (the same workers moving on), or busy at the same moments. We let the seed choose the test, by checking which one best finds seed members we pretend not to know.
\end{plainbox}

\begin{table*}[t]\centering\small
\caption{Forensic expansion among all keys. Repeat run: 703 keys, post hoc (20 random seeds per cell). Two-replica run: frozen, 742 keys, balanced / affinity phase. AUC unless marked; prec.: precision at pool size.}\label{tab:forensic}
\setlength{\tabcolsep}{5pt}
\begin{adjustbox}{max width=\textwidth}
\begin{tabular}{@{}lcccccc@{}}
\toprule
 & \multicolumn{3}{c}{Repeat run, seed share} & Repeat, 20\% & Repeat, 20\% & Two-replica, 20\%\\
Pool & 10\% & 20\% & 30\% & 10 / 20 min & prec. & 20 min, bal.\ / aff.\\
\midrule
shuffle (P1 / F7) & 0.92 & 0.97 & 1.00 & 0.86 / 0.94 & 0.72 & 0.95 / 0.92\\
company C & 1.00 & 1.00 & 1.00 & 0.98 / 1.00 & 1.00 & 1.00 / 1.00\\
per-worker (P4) & 0.67 & 0.90 & 0.96 & 0.73 / 0.85 & 0.59 & --\\
least-busy (P2) & 0.63 & 0.76 & 0.93 & 0.58 / 0.62 & 0.29 & --\\
agents (P6) & 0.5 & 0.5 & 0.5 & -- & -- & --\\
consistent hash (F3) & -- & -- & -- & -- & -- & 0.94 / 0.94\\
hedged (F5) & -- & -- & -- & -- & -- & 0.98 / 0.98\\
rotate on 429 (F1 / F2) & -- & -- & -- & -- & -- & 0.79--0.81\\
sticky, 60\,s (F4) & -- & -- & -- & -- & -- & 0.74 / 0.64\\
shared ring (P3 / F6) & -- & -- & -- & -- & -- & 0.72 / 0.72\\
\bottomrule
\end{tabular}
\end{adjustbox}
\end{table*}

\section{Site-wide discovery}\label{sec:discovery}
Expansion needs a seed. A campaign that content never flags, such as a relay or one with covert prompts, provides none. If it persists, though, it can still be found blind.

\textbf{Spectral rings.} We correlate high-pass occupancy across all $N$ keys, remove the diagonal and the site-wide common mode (hollowing~\cite{khilar2026}), and compare the spectrum with that of key-wise time-shifted surrogates, using the surrogate bulk edge multiplied by 1.05.
\begin{proposition}[Discoverability]\label{prop:bbp}
Null entries have variance $\sigma^2\approx\tau_c/T$. A ring of strength $\theta_1$ produces an outlier eigen-pair when $\theta_1>\kappa\sigma\sqrt N$, that is, when $T>\kappa^2N\tau_c/\theta_1^2$, with $\kappa\ge1$ for heterogeneous activity.
\end{proposition}
\begin{plainbox}\plainlead
$\tau_c$ is how long a key's activity ``remembers'' itself, a few seconds. A window of length $T$ contains roughly $T/\tau_c$ independent snapshots, so chance correlations between unrelated keys shrink like $\sqrt{\tau_c/T}$. Among $N$ keys, these small chance values add up to a noise floor near $\sigma\sqrt N$. The ring is found once $\theta_1$ rises above that floor. Twice as many keys need twice as long, and a ring twice as strong needs a quarter of the time. $\kappa$ is a correction for the fact that real keys are far from identical, and we measure it rather than derive it.
\end{plainbox}
With the measured $\kappa\approx1.9$ and $\tau_c=1.56$\,s, a strong ring among $10^5$ keys needs about 3.4\,h of data, and the jittered ring about 21\,h. That is slower than a two-hour content ban, so blind discovery is for campaigns that content does not see and that persist. Version~0.5 selects keys that load above $2.5/\sqrt N$ on any outlier and splits them into rings by pairwise correlation above $\max(0.2,6\sigma)$, which handles the harmonics of Corollary~\ref{cor:harm}. Version~0.6 takes the union over 20, 60 and 180\,s blocks. Multiplying a vector by the hollowed matrix needs only sparse passes over the occupancy matrix, so $k$ outliers cost $O(k\,\mathrm{nnz})$ by Lanczos iteration (Appendix~\ref{app:lanczos}).

Keys that one worker holds for many requests have long positive autocorrelation. They inflate the null bulk edge and hide rings (Section~\ref{sec:sitewide}). Version~0.8 therefore first drops keys whose \emph{persistence time} is a robust outlier, more than 5 robust standard deviations above the site median. Persistence time is the high-pass autocorrelation integrated up to its first zero crossing. A ring key is visited periodically, so its autocorrelation oscillates and its persistence stays short; a held key's does not. Version~0.7 used instead the integrated \emph{squared} autocorrelation over 120\,s. That counts every periodic peak, and in the rehearsal of the Two-replica run it trimmed half of a ring. Both rules use no labels.
\begin{plainbox}\plainlead
A key that one worker keeps busy for a long stretch has a long memory in time, and many such keys raise the noise floor. We remove them first. The trick is to measure stickiness in a way that does not also remove ring keys, which are busy in regular bursts rather than continuously.
\end{plainbox}

\textbf{Sideband hand-off graph.} For every pair of keys we count end$\to$start hand-offs within 1\,s. A Poisson screen with Benjamini--Hochberg control ($q=10^{-3}$) proposes edges. Each proposed edge is then retested against the same keys' own starts in a $\pm60$\,s sideband, which cancels the effect of bursty sessions. This v0.4 graph has two weaknesses, which the rotate-on-429 pools exposed in the rehearsal of the Two-replica run (Appendix~\ref{app:posthoc}). A key held until it is refused ends its requests in bursts, and the sideband then counts one start after a burst as several hand-offs. And the retest corrects only over the screened edges, so it reuses the counts that selected them. The v0.8 graph counts \emph{distinct} starts that follow some end, takes as its null the share of the local sideband covered by the other key's end windows, and corrects over all pairs of keys.

\section{Active timing watermark}\label{sec:wm}
For every request the provider draws a secret fair coin $b_r$ and delays the end of the stream by $b_r\delta$. Let $L_r$ be the mean lag, measured from $c_r$, of starts on \emph{other} keys of a candidate set within $[c_r,c_r+w)$, and let $\hat T=\overline L_{b=1}-\overline L_{b=0}$.
\begin{proposition}[Exact test, echo, power]\label{prop:wm}
(i) If other keys' starts do not depend on the coins, re-drawing the coins gives an exact p-value~\cite{fisher1935}. (ii) If a worker's next request goes to another key with probability $q$, a fraction $F$ of gaps fall inside the window, and other starts arrive at rate $\lambda_{\rm bg}$, then $\E\hat T\approx\pi\delta$, with $\pi=qF/(qF+\lambda_{\rm bg}w)$. (iii) For $n$ completions, $z\approx\pi\delta\sqrt n/(2\sigma_L)$, with $\sigma_L\approx w/\sqrt{12(1+\lambda_{\rm bg}w)}$.
\end{proposition}
\begin{plainbox}\plainlead
(i) The coins are ours, so under the hypothesis that the keys are unrelated nothing in the traffic depends on them, and every reshuffling of the coins is equally likely. Counting how many reshufflings give an effect as large as the real one gives a p-value that is exact, not approximate. (ii) Only some of the starts in the window come from the same worker; the rest come from everyone else. The echo is the delay $\delta$ diluted by that share $\pi$. (iii) The echo itself stays the same size, but averaging over more answers shrinks the noise around it, so the evidence ($z$) grows like the square root of the number of answers observed.
\end{plainbox}
\emph{Hold-back} delays only the final chunk of the answer, whereas \emph{spread} adds $\delta/n$ to each of the $n$ tokens. Linking two disjoint sets of keys uses ends on one set and starts on the other. The same-key version detects automation behind one key, not linkage.

\subsection{Adversarial analysis}\label{sec:wmadv}
\textbf{Visibility.} Hold-back leaves a 150\,ms gap before \texttt{[DONE]}, against median inter-chunk gaps of 24--27\,ms. A client can read the coin from that one gap with AUC 1.00 (Stress run), subtract the delay before dispatching, and so remove the echo. Spreading left the final gap at 29.6\,ms against 28.1\,ms, and the coin readable with AUC 0.55 (Evasion run). An attacker who cannot read individual coins cannot cancel them. With a clean reference it can still tell that a watermark exists: at AUC 0.55, a rank test needs about 600 completions per group to reach $z=3$. Drawing $\delta$ from a continuous distribution would remove the two-point structure that this test exploits.

\textbf{Jitter.} Uniform client jitter of $J$ seconds pushes the echo out of the window and fills the window with background starts. Proposition~\ref{prop:wm}(ii)--(iii) predicts how long one must observe to reach $z=4$ with the best window. At $\delta=150$\,ms the time needed is 0.2\,h for 1\,s of jitter, 0.65\,h for 2\,s, 4.4\,h for 6\,s and 19\,h for 15\,s. At $\delta=500$\,ms it is 0.12, 0.86 and 4.2\,h for 2, 6 and 15\,s. With a fixed 1\,s window over 36 minutes, the echo vanished at 2\,s of jitter ($p=0.80$). The model reproduced the 6\,s case ($z=1.33$ predicted, 1.26 observed). Jitter costs the attacker 25\%, 49\% and 72\% of per-worker throughput at 2, 6 and 15\,s, which it can win back by adding workers. Like the passive tests, the watermark catches default tooling. Unlike the ring statistic, it does not survive cheap jitter.

\textbf{Buffering.} A client that buffers whole answers and dispatches on a timer cuts the link between completion and dispatch. That is an open loop, covered by Proposition~\ref{prop:imposs}.

\textbf{Cost.} A human reader is not delayed by a held final chunk, because the text has already appeared on screen. A script waiting for the stream to close is delayed by the full $\delta$. Spreading adds about 1.5\,ms per token. The measured cost for all users at $\delta=150$\,ms was $+76$\,ms on average. Under stress that is $+1.2\%$ at the median answer time (5.69$\to$5.76\,s) and $+0.4\%$ at the 99th percentile (28.47$\to$28.57\,s). We recommend applying it only to keys already under suspicion. If those keys are chosen from earlier data, before the coins are drawn, the test stays exact given the selection.

\section{Limits and the cost of evasion}\label{sec:limits}
\begin{proposition}[Indistinguishable dispatchers]\label{prop:imposs}
(i) $K$ independent workers pinned to their own keys have the same joint distribution as $K$ independent single-key batch clients. (ii) Open-loop Poisson arrivals on random keys are independent Poisson streams. Against a benign population that contains such clients, every level-$\alpha$ timing test, the watermark included, has power at most $\alpha$.
\end{proposition}
\begin{plainbox}\plainlead
If the campaign's traffic is statistically identical to honest traffic, then any test that rarely accuses honest users ($\alpha$ of the time) accuses the campaign just as rarely. No cleverness can get around this, because there is simply nothing in the timestamps to find.
\end{plainbox}
This is the timing form of the law-matching barrier of~\cite[Prop.~1]{khilar2026}. An open-loop sender under a per-key cap of $Q$ requests per minute must give up about $z/\sqrt Q$ of its quota to keep overruns rare: 26\%, 9\% and 3.7\% at $Q=60$, 500 and 3{,}000 for $z=2$. Pinning costs nothing in throughput when each key's quota covers one worker. When the quota binds, the attacker's throughput is at most the number of keys times the quota, whatever the dispatcher. In the Two-replica run, under a 15/min quota (a cap of 0.25 requests/s per key), pinned workers ran at 0.239 requests/s each, against 0.30 for workers rotating over 40 keys. A pinned campaign recovers that throughput by adding keys and workers, not by changing its dispatcher. Its cost lies elsewhere: pinning concentrates the campaign's queries on few keys.

\begin{proposition}[Content--timing trade-off]\label{prop:pincer}
A campaign sends $n$ queries through $M$ keys with at most $K$ concurrent workers. Suppose a stateful per-key content monitor flags a key that has carried $v$ queries at $z\approx\epsilon\sqrt v$, for a per-query signal $\epsilon>0$. (i) Keeping every key below a content threshold $z^*$ needs $M\ge M^*=n\epsilon^2/z^{*2}$ keys. (ii) If $M>K$, the campaign makes at least $M-K$ \emph{switches}, where a switch is a worker's first request on a key it did not start on. In a closed loop each switch is an end$\to$start hand-off between two of its keys. (iii) Hence a campaign that evades the content monitor makes at least $n\epsilon^2/z^{*2}-K$ switches. It makes none only if $K\ge M^*$, that is, if it runs one concurrent, pinned worker per key.
\end{proposition}
\begin{plainbox}\plainlead
A content monitor grows more confident the more queries a key carries, so a campaign must spread its queries over many keys to stay below the alarm. If it has fewer workers than keys, some worker must at some point move from one key to another, and in a closed loop that move leaves a hand-off in the timing, though a single hand-off is weak evidence. The only way to avoid every move is to run one worker per key, all at once.
\end{plainbox}
The trade-off is real but narrow. Workers are coroutines, so an attacker that can afford $M^*$ live identities at once escapes both channels, and Proposition~\ref{prop:imposs} covers that case. The proposition therefore bites only when concurrency is what is scarce: per-organisation concurrency limits, one exit address per live worker, or accounts that must look human-paced. It guarantees that switches exist, not that each one can be detected. A switch that retires one key for good is a single hand-off, and its evidence is weak unless switches recur. The documented proxy networks that ran 20{,}000 accounts at once~\cite{anthropic2026} show that some attackers do pay for concurrency, and they relayed real users, which is the open-loop case of Proposition~\ref{prop:imposs}(ii).

\section{Testbed and protocol}\label{sec:testbed}
Qwen2.5-0.5B-Instruct~\cite{qwen2024} runs on vLLM~\cite{kwon2023} 0.32.0 on a single T4 GPU. In front of it sits an aiohttp gateway that logs, for every request, the arrival, first byte, model completion, last byte, status, bytes sent and the watermark coin. Each population of clients runs in its own process and tags its requests with IDs, so client-side and gateway-side times can be joined (median delay 11\,ms).

\textbf{Runs.} We refer to each experiment by a short name, summarised in Table~\ref{tab:runs}. The \emph{Pilot} run was our first live check, with every client population sharing one event loop. The \emph{Baseline} run tested all the dispatchers with short, 96-token answers over 40 minutes, about 2.3\,s per answer. The \emph{Repeat} run is a frozen re-test of the Baseline, with a new random seed and the hold-back watermark. The \emph{Stress} run is a frozen re-test on a struggling server. Its prompt and output lengths come from WildChat (median output 156 tokens, 90th percentile 619), and its simulated humans use WildChat think times (median 74\,s) and session lengths. We limited the KV cache, the server's memory for answers in progress, to 1{,}600 blocks, and usage reached 0.999 at the 90th percentile, with 48{,}789 preemptions (requests paused to free memory), up to 65 queued requests and answers of 6\,s. The \emph{Evasion} run is our sweep of evasion strategies on 20-key pools, with the watermark spread over the stream. The \emph{Two-replica} run put two vLLM replicas on one GPU behind the gateway, enforced a quota of 15 requests per minute per key (HTTP 429 beyond it), and added the F pools of Table~\ref{tab:pops}. In its \emph{balanced} phase each request went to the replica with fewer outstanding requests, and in its \emph{affinity} phase each key was pinned to one replica by hash. Each phase lasted 40 minutes. Before every live run we rehearsed it in simulation, and the \emph{rehearsal} of the Two-replica run comes up several times below because it caught two detector problems before the live run. The \emph{WildChat} study (Section~\ref{sec:wildchat}) and the \emph{Production} study (Section~\ref{sec:chutes}) use real traffic.

\begin{table}[t]\centering\small
\caption{The experiments, by name.}\label{tab:runs}
\setlength{\tabcolsep}{3pt}
\begin{adjustbox}{max width=\columnwidth}
\begin{tabular}{@{}lp{5.4cm}@{}}
\toprule
Name & What it tests\\
\midrule
WildChat & false alarms among 1.96 million real requests (imputed starts)\\
Pilot & first live check; all clients on one event loop\\
Baseline & every dispatcher, short answers, 40 minutes\\
Repeat & frozen re-test of the Baseline; hold-back watermark\\
Stress & frozen re-test with real answer lengths and a saturated server\\
Evasion & jitter, held keys, pinning and open loops; spread watermark\\
Two-replica & two replicas, per-key quotas, five unseen dispatchers; balanced and affinity phases\\
Production & three hours of a public serving trace, two held out; injected and real pools\\
\bottomrule
\end{tabular}
\end{adjustbox}
\end{table}

\textbf{Protocol.} Every experiment locked its predictions before running, and we report each outcome (Appendix~\ref{app:ledger}). Every detector version was fixed before the live run that tested it, using simulation or earlier runs only. Main tables report frozen results, and post-hoc analyses are labelled. We handle multiple testing the same way throughout: within each run, all watermark p-values for non-human pools (cross-key, halves and same-key) form one Holm family at $\alpha=0.05$. Human pools serve as calibration checks and are reported as counts below $p=0.01$.

\section{Results}\label{sec:results}
\subsection{Testbed artefacts}
\begin{figure}[t]\centering
\includegraphics[width=\columnwidth]{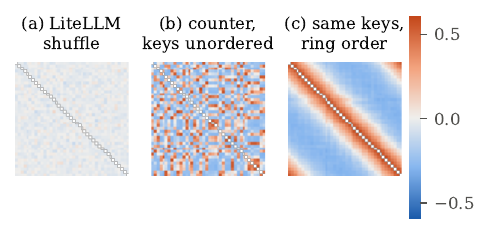}
\caption{High-pass occupancy correlation between the 40 keys of a pool (Baseline run). (a) LiteLLM shuffle: uniform weak turn-taking. (b) Shared counter with keys in arbitrary order. (c) The same keys ordered by circular seriation of (b) alone: the travelling wave of Proposition~\ref{prop:ring}.}\label{fig:heat}
\end{figure}

In the Pilot run the LiteLLM shuffle pool gave $R=0.412$, against $0.41$ predicted from its own Fano factor ($z=22.7$). Contiguous halves of a shared ring were bunched ($R_{20}=2.58$, $z=-59$), while random halves took turns. That contrast is the travelling wave seen directly: neighbouring keys are busy together, while keys picked at random from the ring are not (Fig.~\ref{fig:heat} shows the same wave in the Baseline run). The shared event loop also produced three artefacts. The random relay showed $R=0.85$, batch jobs looked bunched ($R_{20}=1.24$), and the Fano count overshot (13.7 for 8 workers). All three disappeared once each population ran in its own process, which is why every later run does so.

\subsection{Forensic expansion}
From a 20\% seed (Table~\ref{tab:forensic}), the rest of the shuffle pool ranked at AUC 0.97 among 703 keys, and the company at 1.00. Per-worker rotations reached 0.90 and least-busy 0.76. Agents, which hold one key per episode, could not be expanded at all. A seed that was half impure cost 0.05--0.15 of AUC. Ten minutes of data already gave 0.86 for shuffle and 0.98 for the company, inside the window between a first alarm and a ban. Precision at pool size was 0.72 for shuffle and 1.00 for the company.

The Repeat-run analysis was post hoc, so the Two-replica run froze it as prediction M8 (AUC $\ge0.85$ within 20 minutes), on two replicas with quotas and 742 keys. From a 20\% seed, shuffle, company, consistent hashing and hedged requests ranked at AUC 0.95, 1.00, 0.94 and 0.98 in the balanced phase, and 0.92, 1.00, 0.94 and 0.98 in the affinity phase. After 10 minutes the AUC was already 0.82--0.98. Pools that hold a key until it is refused or times out expanded less well (0.64--0.81). Rings expand poorly from a seed, because their correlation is local to neighbouring keys, but blind discovery finds them whole.

\subsection{Dispatcher identification, frozen}
In the Repeat run every pool was classified correctly (Table~\ref{tab:pools}). In the Stress run, v0.5 and v0.6 both classified 11 of 12 correctly; the miss was the random relay, flagged as bunching for reasons explained in Section~\ref{sec:crosstalk}. The three benign pools (humans, humans with batch jobs, pinned keys) were classified as none in both runs. The legitimate company (C) is classified exactly like a shuffling campaign, and Section~\ref{sec:deploy} discusses what that means in practice.

The worker count came within 0.4 of the truth for all ten always-busy pools across both runs, and the Fano count was exact. Under stress, over a longer window, the rings were at least as strong as before, at ring $z=42$--$51$, even though they now took 28--58\,s to go round. We had revised our prediction to say that such slow rings would be lost at the 20\,s scale, and that was wrong. The correlation comes from overlapping answers on neighbouring keys, so it lives on the timescale of a single answer, not of a full lap. LiteLLM \texttt{least-busy} was classified as sticky in both runs, and the watermark shows why (Section~\ref{sec:wmres}).

\begin{table*}[t]\centering\small
\caption{Frozen re-tests: the Repeat run (clean, 96-token answers) and the Stress run (WildChat lengths, saturated KV cache, 48{,}789 preemptions). $z_R$ is against 200 random 40-key human pools. Ring $z$ is the maximum over scales (v0.6), against 40 time-shifted surrogates per scale; in the Repeat run it is at the 20\,s scale only. Class: frozen v0.5 = v0.6 except where marked. $K$: point estimate (truth).}\label{tab:pools}
\setlength{\tabcolsep}{3.5pt}
\begin{adjustbox}{max width=\textwidth}
\begin{tabular}{@{}l|cccccl|cccccl@{}}
\toprule
 & \multicolumn{6}{c|}{Repeat run (clean)} & \multicolumn{6}{c}{Stress run}\\
Pool & $R$ & $z_R$ & ring $z$ & succ.\ $z$ & $K$ (true) & class & $R$ & $z_R$ & ring $z$ & succ.\ $z$ & $K$ (true) & class\\
\midrule
P1 shuffle & 0.096 & 15.9 & 2.8 & $-13.0$ & 6.0 (6) & shuffle & 0.033 & 15.8 & 1.8 & $-12.0$ & 6.3 (6) & shuffle\\
P2 least-busy & 0.441 & 9.9 & 2.8 & $-2.1$ & 9.9 (10) & sticky & 0.431 & 9.5 & 1.6 & $-1.5$ & 10.4 (10) & sticky\\
P3 shared counter & 0.122 & 15.4 & 47.9 & $-5.0$ & 12.1 (12) & shared, closed & 0.041 & 15.7 & 51.3 & $-5.6$ & 12.1 (12) & shared, closed\\
P4 per-worker & 0.089 & 16.0 & 1.2 & 79.5 & 8.1 (8) & per-worker & 0.030 & 15.9 & 2.4 & 70.4 & 8.2 (8) & per-worker\\
P5 counter + jitter & 0.636 & 6.5 & 24.2 & $-6.4$ & 9.1 (8) & shared, closed & 0.356 & 10.7 & 42.2 & $-4.6$ & 8.3 (8) & shared, closed\\
P6 agents & 0.601 & 7.1 & 0.5 & 2.8 & 5.2 (6) & sticky & 0.549 & 7.7 & 1.5 & 2.3 & 6.2 (6) & sticky\\
P7 relay, random & 1.068 & $-1.0$ & 0.8 & $-1.7$ & -- & none & 1.364 & $-5.1$ & 0.0 & 0.4 & -- & \textbf{bunching}\\
P8 relay, round-robin & 1.173 & $-2.8$ & 50.8 & $-9.2$ & -- & shared, open & 1.564 & $-8.3$ & 51.5 & $-0.8$ & -- & shared, open\\
C company (benign) & 0.097 & 15.8 & $-3.3$ & $-7.3$ & 4.0 (4) & shuffle & 0.035 & 15.8 & 0.1 & $-4.7$ & 4.4 (4) & shuffle\\
P9 pinned & 1.011 & 0.0 & 1.0 & $-0.2$ & 8.2 (8) & none & 1.003 & 0.5 & $-0.5$ & $-1.7$ & 8.2 (8) & none\\
humans (40 random) & 1.112 & $-1.8$ & 0.9 & 0.1 & -- & none & 1.050 & $-0.2$ & 1.4 & -- & -- & none\\
36 humans + 4 batch & 0.992 & 0.3 & $-1.3$ & 0.4 & -- & none & 1.069 & $-0.5$ & 1.3 & 0.5 & -- & none\\
\midrule
correct & \multicolumn{6}{c|}{12 / 12} & \multicolumn{6}{c}{11 / 12 (v0.5 and v0.6)}\\
\bottomrule
\end{tabular}
\end{adjustbox}
\end{table*}

\subsection{Replicas, quotas and unseen dispatchers}\label{sec:nb9}
The Two-replica run used 742 keys on two vLLM replicas with a quota of 15 requests per minute per key. Its balanced phase spread requests across replicas by outstanding load, and its affinity phase pinned each key to a replica (Table~\ref{tab:nb9}).

\textbf{Provider-side load balancing did not create false turn-taking.} Random 40-key human pools stayed at the null, with $R=1.004\pm0.046$ and $0.998\pm0.045$, and only 0.5\% and 1\% of pools beyond $\pm3$\,sd. The human pool's ring $z$ was 0.5 and $-0.2$. In the affinity phase, pools drawn from the humans of a single replica had mean $z=0.25$ and $-0.22$. Decode time per byte did depend on load (Spearman 0.30 and 0.15), so batching coupled durations as it had in the Stress and Evasion runs, but that coupling did not reach the occupancy ratio.

\textbf{Unseen dispatchers.} All five new dispatchers were flagged in both phases, and the two with a predicted class received it: hashing came out as shuffle, and sticky-with-limit as sticky. Rotation on 429 with a \emph{shared} pointer was classified as per-worker rotation. In hindsight this makes sense: all workers use one key until it is refused and then move together to the next, so each key's successor is fixed (successor $z=25$ and 44). This pool also showed the strongest turn-taking we have measured ($R=0.014$). Hedged requests were classified as shuffle. With two streams per worker, the Fano count found the 12 streams in flight, and the worker count gave 9.2 and 9.4 for 6 workers, because a cancelled hedge ends early. Every other always-busy pool was counted to within 0.2 workers, and the reference pools and the company kept their earlier classes.

\textbf{Failover after refusal.} After a 429, another key of the same pool started within 0.3\,s far more often than its local rate would predict. The excess was $z=20$ and 21 for the shared pointer, 53 and 55 for per-worker pointers, and 24 and 22 for sticky keys, which also move on when refused. Pinned workers retry their own key, so other keys started \emph{less} often than expected ($z=-4.7$, $-4.8$). This gives one more signature, and it needs no model of the dispatcher, because the provider itself issued the refusals.

\textbf{Site-wide.} Spectral v0.5, v0.6 and v0.8 recovered the shared ring at 40/40 in both phases, pure, in exact ring order, and with no humans. v0.7 trimmed 73 and 80 keys and recovered none of the ring, as we had predicted after the rehearsal. Both hand-off graphs recovered the two rotate-on-429 pools and the ring at 100\%. Their human selections are the Two-replica run's one failure. The frozen v0.4 graph selected 24 and 23 humans, all linked to rotate-on-429 pools, as the rehearsal had predicted (M9). The v0.8 graph, built to remove them, still selected 2 and 4, above the locked limit of 2 in the affinity phase. Spectral discovery selected no human in either phase. Its null uses random surrogates, so we re-ran v0.8 from the released logs with 10 surrogate seeds per phase. The ring was recovered at 40/40 every time. The balanced phase never selected a human. The affinity phase selected one human with 2 of the 10 seeds, and with 5 of them it also selected keys of the rotate-on-429 and sticky pools, which are campaign keys.

\begin{table*}[t]\centering\small
\caption{Two-replica run, frozen: two replicas, 15/min quota per key, balanced phase (load balancing) and affinity phase (key affinity). Notation as in Table~\ref{tab:pools}; ring $z$ is multi-scale; class by v0.7 (identical for v0.6).}\label{tab:nb9}
\setlength{\tabcolsep}{3.5pt}
\begin{adjustbox}{max width=\textwidth}
\begin{tabular}{@{}l|cccccl|cccccl@{}}
\toprule
 & \multicolumn{6}{c|}{Balanced phase (least outstanding)} & \multicolumn{6}{c}{Affinity phase (key pinned to a replica)}\\
Pool & $R$ & $z_R$ & ring $z$ & succ.\ $z$ & $K$ (true) & class & $R$ & $z_R$ & ring $z$ & succ.\ $z$ & $K$ (true) & class\\
\midrule
F1 429, shared ptr. & 0.014 & 21.8 & 0.7 & 24.9 & 7.9 (8) & per-worker & 0.013 & 21.9 & 2.1 & 44.1 & 8.0 (8) & per-worker\\
F2 429, per worker & 0.017 & 21.7 & 1.9 & 26.3 & 8.0 (8) & per-worker & 0.015 & 21.9 & $-2.9$ & 26.6 & 8.1 (8) & per-worker\\
F3 consistent hash & 0.075 & 20.4 & 1.2 & $-11.8$ & 8.1 (8) & shuffle & 0.076 & 20.5 & 0.0 & $-11.6$ & 7.9 (8) & shuffle\\
F4 sticky, 60\,s & 0.084 & 20.2 & 2.0 & 0.4 & 8.2 (8) & sticky & 0.082 & 20.4 & $-0.4$ & 0.9 & 7.9 (8) & sticky\\
F5 hedged & 0.307 & 15.3 & 0.3 & $-16.9$ & 9.2 (6) & shuffle & 0.288 & 15.8 & 2.7 & $-14.2$ & 9.4 (6) & shuffle\\
F6 shared counter & 0.092 & 20.1 & 19.9 & $-2.1$ & 7.9 (8) & shared, closed & 0.090 & 20.2 & 30.8 & $-2.6$ & 8.1 (8) & shared, closed\\
F7 shuffle & 0.075 & 20.5 & 0.6 & $-12.8$ & 7.9 (8) & shuffle & 0.076 & 20.5 & 0.4 & $-10.4$ & 8.0 (8) & shuffle\\
F8 pinned & 1.045 & $-0.9$ & $-0.3$ & 0.4 & 6.3 (8) & none & 1.018 & $-0.4$ & 1.2 & $-0.6$ & 6.2 (8) & none\\
F9 open loop, steady & 1.067 & $-1.4$ & 1.1 & $-0.8$ & -- & none & 1.078 & $-1.8$ & 0.1 & $-0.2$ & -- & none\\
C company (benign) & 0.087 & 20.2 & $-0.5$ & $-5.7$ & 4.1 (4) & shuffle & 0.085 & 20.3 & $-1.2$ & $-6.8$ & 4.1 (4) & shuffle\\
humans (40 random) & 1.017 & $-0.3$ & 0.5 & $-1.2$ & -- & none & 0.893 & 2.4 & $-0.2$ & 0.5 & -- & none\\
36 humans + 4 batch & 1.003 & 0.0 & 0.7 & $-1.5$ & -- & none & 1.059 & $-1.3$ & 0.9 & 0.8 & -- & none\\
\midrule
correct & \multicolumn{6}{c|}{12 / 12} & \multicolumn{6}{c}{12 / 12}\\
\bottomrule
\end{tabular}
\end{adjustbox}
\end{table*}

\subsection{The theory on live data}
\begin{figure}[t]\centering
\includegraphics[width=\columnwidth]{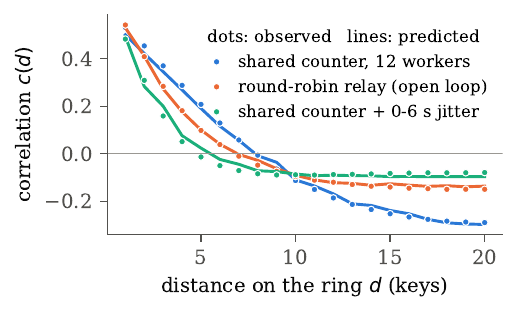}
\caption{Correlation of keys at ring distance $d$ (Baseline run). Dots: observed. Lines: Proposition~\ref{prop:ring} from the pool's own visit delays and durations, with no fitted parameter.}\label{fig:profile}
\end{figure}

Fig.~\ref{fig:profile} compares the measured ring profiles with Proposition~\ref{prop:ring}, computed from each pool's own durations and visit delays. The root-mean-square error over $d\le20$ was 0.015--0.020, and the predicted $\theta_1$ matched each pool's top eigenvalue: 6.83/6.91, 4.58/4.67 and 2.73/2.79. Corollary~\ref{cor:harm} also explained two Repeat-run outliers that v0.4 had left unexplained. The jittered ring ($\lambda\E D=3.1$ keys) has $\theta_{1,2}=(2.88,2.34)$, the relay ($4.6$ keys) $(4.86,2.30)$, and the shared counter ($9.5$ keys) $(7.09,1.81)$. Adding the BBP shift, the second harmonics of the jittered ring and the relay land at the observed pairs 3.35/3.22 and 3.01/2.94.

\subsection{Site-wide discovery}\label{sec:sitewide}
\begin{figure}[t]\centering
\includegraphics[width=\columnwidth]{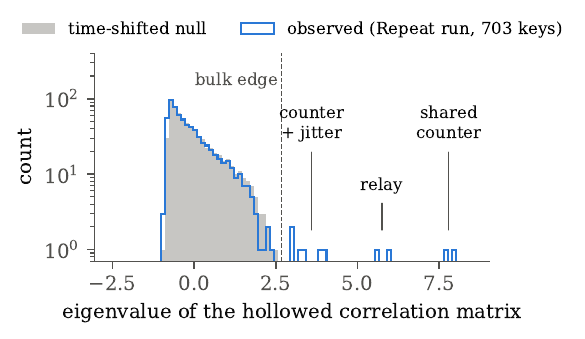}
\caption{Eigenvalues of the hollowed key-by-key correlation matrix over the 703 keys of the Repeat run, against a time-shifted surrogate. Each outlier is a degenerate pair belonging to a rotation ring; the counter + jitter ring has two (Corollary~\ref{cor:harm}).}\label{fig:spec}
\end{figure}

\textbf{Clean conditions (Repeat run).} With the frozen v0.4, spectral discovery recovered the shared ring at 100\% and the relay ring at 98\%, all pure, with zero humans and the full ring order. It split the jittered ring (75\%) across two harmonics (Fig.~\ref{fig:spec}). The sideband hand-off graph, which we had fixed after the Baseline run pulled in 17 humans through overlapping sessions, recovered the per-worker and shared rotations at 100\% with zero humans.

\textbf{Under stress (Stress run).} Frozen v0.5 recovered the shared ring and the relay at 100\% but missed the jittered ring. Frozen v0.6 recovered all three at 100\%, pure, with zero humans. The hand-off graph recovered the per-worker ring and the company at 100\% but only 35\% of the shared ring, which tells us that long answers weaken hand-offs.

\textbf{Evasion sweep (Evasion run).} v0.5 found four of the 20-key rings (no jitter, 2\,s jitter, 24 workers, key held for 5 requests) at 100\%, pure, with zero humans. v0.6 found only one, because keys held for many requests inflated the null bulk edge from 2.6 to 4.6 (Appendix~\ref{app:posthoc}). Trimming those keys without labels recovered all five shared-counter rings at 20/20, pure. Our first trimming rule, by correlation time (v0.7), then failed its first test, on the simulated rehearsal of the Two-replica run: the ring's own keys had the longest correlation times, and v0.7 removed 20 of its 40 keys. Trimming by persistence time (v0.8) keeps all five Evasion-run rings at 20/20, trims exactly the held-key pool, and recovered the rehearsal ring. On the live Two-replica run, frozen v0.8 recovered the ring at 40/40 in both phases, pure, with no humans, and v0.7 recovered none of it.

\subsection{The watermark}\label{sec:wmres}
Table~\ref{tab:wm} summarises both watermark runs. With hold-back, every closed-loop pool that reassigns keys echoed at $z=12.6$--$17.6$. Disjoint halves of each pool were linked at the permutation floor, both in the Repeat run and in the Stress run, and all of these survive Holm correction. The human pools were well calibrated, with 0 of 50 below $p=0.01$ in both runs. The spread design kept every jitter-free pool that moves to a new key on each request ($z=11$--$13$). Pools that hold a key for several requests echo on the same key instead. From the measured rates alone, Proposition~\ref{prop:wm} predicted the size of the echo to within 14\%.

Two results changed under uniform Holm correction. The halves test for the jittered pool in the Repeat run ($p=0.0095$) is not significant. The random relay's cross-key echo in the Repeat run ($p=0.0015$) is significant within that run's family of 24 tests, but it did not replicate in the Stress or Evasion runs (cross-key $z=0.9$ in both), so we treat it as unexplained. The coins also corrected one of our own conclusions. LiteLLM \texttt{least-busy} echoes almost only on the \emph{same} key: about 85\% of its follow-up requests return there. The router is mostly pinned, and its occasional migrations are what its long-timescale $R$ picks up.

\begin{table*}[t]\centering\small
\caption{Watermark ($\delta=150$\,ms, window 1\,s; $p=0.0005$ is the floor). Each cell: cross-key $z$ / halves $p$ / same-key $z$.}\label{tab:wm}
\setlength{\tabcolsep}{5pt}
\begin{adjustbox}{max width=\textwidth}
\begin{tabular}{@{}lccc@{}}
\toprule
Pool & Repeat, hold-back & Stress, hold-back & Evasion, spread\\
\midrule
shuffle (P1 / E7) & 14.2 / .0005 / 0.8 & 16.0 / .0005 / 1.2 & 11.3 / .0005 / 1.6\\
shared counter (P3 / E0) & 13.2 / .0005 / -- & 16.5 / .0005 / -- & 13.1 / .0005 / 0.9\\
per-worker (P4) & 13.0 / .0005 / 0.2 & 17.6 / .0005 / $-0.8$ & --\\
company C & 12.6 / .0005 / 4.3 & 13.9 / .0005 / 2.9 & --\\
least-busy P2 & $-1.4$ / .65 / \textbf{28.6} & $-0.1$ / .56 / \textbf{26.6} & --\\
pinned (P9 / E9) & $-1.9$ / .52 / \textbf{43.9} & $-0.2$ / .22 / \textbf{30.0} & $-0.2$ / .70 / \textbf{34.4}\\
jitter 6\,s (P5 / E2) & 1.3 / .0095 / -- & 1.0 / .47 / -- & 0.5 / .53 / --\\
relay, random (P7 / E10) & 2.9 / .60 / 1.4 & 0.9 / .35 / 0.1 & 0.9 / .17 / 0.8\\
relay, round-robin (P8) & $-0.5$ / .15 / -- & $-0.3$ / .24 / -- & --\\
\midrule
humans: \# $p<0.01$ of 50 & 0 & 0 & 0 (of 46)\\
client sees coin (AUC) & -- & 1.00 & 0.55\\
\bottomrule
\end{tabular}
\end{adjustbox}
\end{table*}

\subsection{The cost of evasion}\label{sec:cost}
\begin{figure}[t]\centering
\includegraphics[width=\columnwidth]{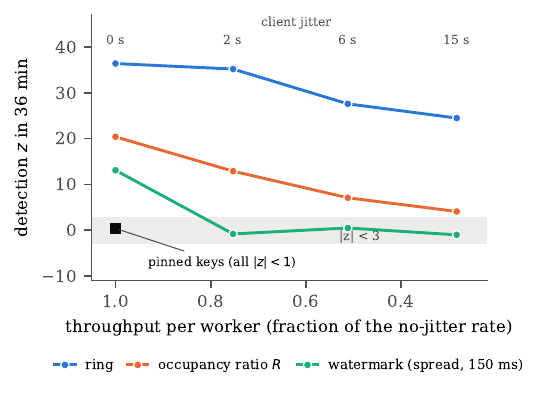}
\caption{Cost of evasion (Evasion run, shared counter over 20 keys, 8 workers). Each point is one jitter level; detection $z$ over 36 minutes against the throughput each worker keeps.}\label{fig:cost}
\end{figure}

Fig.~\ref{fig:cost} and Table~\ref{tab:cost} summarise the Evasion run. The ring survived every evasion that kept rotating keys. It held under jitter of up to 15\,s, which cost 72\% of each worker's throughput (ring $z=24.5$), and it held when the workers were tripled to win that throughput back ($z=29.9$). In that last case occupancy gave $R\approx1$, so only the ring statistic saw the pool. The ring held for keys kept for 5 requests ($z=6.4$), and with 20 requests $R$ still flagged the pool at $z=20$. The spread watermark lost every pool with 2\,s of jitter or more (Section~\ref{sec:wmadv}). Only pinned keys and the steady open loop escaped, at full throughput, exactly as Proposition~\ref{prop:imposs} says they must. Section~\ref{sec:adaptive} repeats this sweep for an adaptive attacker on production background and under binding quotas. The bursty open loop was flagged as bunching ($z=-157$), because all of its keys share its on/off rhythm.

\begin{table}[t]\centering\small
\caption{Evasion sweep (Evasion run). Throughput in requests/s per worker; ring $z$ is multi-scale.}\label{tab:cost}
\setlength{\tabcolsep}{2pt}
\begin{adjustbox}{max width=\columnwidth}
\begin{tabular}{@{}lccccl@{}}
\toprule
Variant & thr./worker & $z_R$ & ring $z$ & wm.\ $p$ & class (v0.6)\\
\midrule
E0 counter & 0.328 & 20.4 & 36.4 & .0005 & shared, closed\\
E1 jitter 2\,s & 0.247 & 12.9 & 35.2 & .80 & shared, closed\\
E2 jitter 6\,s & 0.168 & 7.1 & 27.6 & .31 & shared, closed\\
E3 jitter 15\,s & 0.093 & 4.1 & 24.5 & .85 & shared, closed\\
E4 jitter 15\,s, $K{=}24$ & 0.094 & $-2.5$ & 29.9 & .32 & shared, open\\
E5 hold 5 & 0.330 & 20.3 & 6.4 & .11 & shared, closed\\
E6 hold 20 & 0.331 & 20.3 & 4.2 & .34 & sticky\\
E7 shuffle $K{=}8$ & 0.331 & 21.3 & 0.8 & .0005 & shuffle\\
E8 shuffle $K{=}18$ & 0.332 & 21.2 & 1.9 & .0005 & shuffle\\
E9 pinned & 0.329 & $-0.5$ & 0.9 & .59 & none\\
E10 open loop, steady & -- & $-4.3$ & 1.7 & .18 & \textbf{bunching}\\
E11 open loop, bursty & -- & $-156.9$ & 0.3 & .44 & bunching\\
\bottomrule
\end{tabular}
\end{adjustbox}
\end{table}

\subsection{Server cross-talk}\label{sec:crosstalk}
Batching couples the durations of unrelated requests. The Spearman (rank) correlation between decode time per byte and site load was $-0.08$ in the Pilot run, but $0.28$ under preemption (Stress run) and $0.80$ with 99 sequences batched (Evasion run). When the server slows down, every answer in flight lengthens together, which raises $\Var N$ for high-volume pools. That is why the steady open loop and the random relay were flagged as bunching ($R=1.18$ and $1.36$), and why the round-robin relay's $R$ rose to $1.56$. The bias runs only towards $R>1$, so it cannot create false turn-taking, and the human nulls stayed calibrated (0.5\% of pools beyond $\pm3$). The same ratio computed on request \emph{start} counts never uses durations at all. Applied post hoc to the Stress and Evasion runs, it returned the steady open loop and the random relay to $0.98$ and $1.01$, kept the bursty loop bunched ($z=-68$), and still detected the closed loops, though with less power. Version~0.7 therefore confirms any bunching call on start counts. The Two-replica run never exercised that step. Its steady open loop was never flagged as bunching, not even by v0.6 ($z_R=-1.4$ and $-1.8$), because decode speed depended less on load there (Spearman 0.30 and 0.15, against 0.80 in the Evasion run).
\begin{plainbox}\plainlead
When the server is busy, everyone's answers slow down together, so unrelated keys briefly look ``busy at the same time''. That can only push $R$ up, never down. Counting only when requests \emph{start} gives a version of the test that never uses answer durations, so this effect cannot inflate it.
\end{plainbox}

\subsection{Limits in practice}
Pinned keys behaved exactly as Proposition~\ref{prop:imposs} requires in every run: $R=1.00$--$1.02$, ring and successor statistics at the null, no site-wide detector found them, and there was no cross-key echo. The same-key echo ($z=30$--$44$) does reveal automation behind each key, but legitimate batch jobs show it too ($z=30$).

\subsection{Two channels together}
Table~\ref{tab:two} combines timing with the content channel of~\cite{khilar2026}. Timing alone catches the hardest content campaign to detect (content detection 0.04) whenever that campaign rotates keys through a ring. A covert-prompt campaign that also pins its keys, or uses a steady open loop, evades both channels at almost no extra cost in throughput, unless concurrency is scarce (Proposition~\ref{prop:pincer}). The covert end of the frontier is therefore still set by content, at no more than about 42\% of a dataset campaign's gain~\cite{khilar2026}.

\begin{table*}[t]\centering\small
\caption{Content~\cite{khilar2026} and timing on the same campaign (regimes in Section~\ref{sec:setting}).}\label{tab:two}
\setlength{\tabcolsep}{6pt}
\begin{adjustbox}{max width=\textwidth}
\begin{tabular}{@{}llccl@{}}
\toprule
Content (gain) & Dispatch & Content & Timing & Outcome\\
\midrule
Dataset (+10.4) & any & 1.00 & -- & content\\
Half padding (+7.6) & ring & 0.57 & site-wide & timing\\
Half padding (+7.6) & shuffle & 0.57 & from seed & both\\
Tilted (+2.2) & ring & 0.04 & site-wide & \textbf{timing only}\\
Tilted (+2.2) & shuffle & 0.04 & needs seed & evades\\
Written (+4.4) & pinned / steady open & $\approx0$ & none & evades\\
\bottomrule
\end{tabular}
\end{adjustbox}
\end{table*}

\section{Real traffic: WildChat}\label{sec:wildchat}
Our first look at real users came from WildChat-1M~\cite{zhao2024}: 1.96 million requests from 178{,}724 hashed IP addresses over 387 days. User turns carry no timestamps there, so we had to impute request starts. With that caveat, the benign false-alarm rate passed its locked check for 6-hour windows. Random, same-country and same-browser-and-country pools gave $R_{60}=0.982\pm0.037$, $0.987\pm0.045$ and $0.995\pm0.027$, and none fell below 0.85. Two settings failed: the heaviest accounts (3.8\% of pools below 0.85) and 1-hour windows (about 30\% below 0.85). A calibrated scan of 5{,}528 groups showed that the surrogate null was anti-conservative on real traffic. Control groups crossed $p<0.002$ at 2.3 times the nominal rate, and no group survived Benjamini--Hochberg control. Because the starts were imputed, these numbers bound the difficulty from above rather than measure it. As in~\cite{khilar2026}, the lesson is to calibrate against real past windows, never against reshuffled ones. Testing timing detectors end to end needs gateway logs that record both starts and ends, which only providers hold; the next section uses a provider's published logs.

\section{Production traffic: the Chutes trace}\label{sec:chutes}
Everything so far used synthetic users, and WildChat, though real, had imputed starts and no answer ends. This section, the Production study, tests the frozen detectors on production serving logs that record both, with predictions locked in advance and windows held out.

\textbf{Data.} The Chutes trace~\cite{nixon2026} records 6.12 billion requests served by a public LLM inference platform over one year, released under CC BY 4.0. Each request carries an anonymised account, the serving instance, the model, and start and completion times in microseconds. There is no API-key field, so an account plays the role of a key. We took three one-hour windows, chosen mechanically at 20\%, 50\% and 80\% of the sorted list of shards. W0, the 50\% shard, was our development window, and W1 and W2 were held out. At an activity floor of 10 requests in the hour, the windows had 3{,}014, 1{,}611 and 3{,}069 active accounts. The requests that overlap each window number 881{,}167, 669{,}020 and 643{,}220. Median answer durations were 7.4--7.9\,s. Traffic is dominated by a handful of \emph{platform} accounts with up to 410{,}049 requests per hour, and the ten largest accounts carry 77--87\% of each window.

\textbf{Protocol.} We locked predictions R1--R8 and recorded their hash before downloading any data. A first amendment, also written before we saw any data, fixed the estimators. A second amendment came after the W0 pass and before W1 or W2 was opened. It fixed the activity floor and the strata on W0, as the lock allowed: light accounts send 10--30 requests per hour, heavy accounts 31--2{,}000, and platform accounts more than 2{,}000. Platform accounts stay in every site-wide test but are left out of random-pool tests. The amendment also recorded three implementation changes that give identical output, declared R3 (replica coupling) not testable because no instance hosts 40 accounts that use only it, and registered three new predictions, R2b, R6b and R9. We then hashed the code and ran it once on each held-out window.

The trace contains no known campaign, so we inject the positives. Eleven synthetic pools are simulated with each window's own answer durations and a quota of 15 requests per minute per key, and added to the real background with the same seed in every window. The real accounts are the negatives, except where they share dispatch themselves, which is exactly what R8 measures.
\begin{plainbox}\plainlead
We wrote down what we expected before looking, practised on one hour of data, wrote down what we had learned, locked the code, and only then opened two fresh hours. Since nobody knows which real accounts belong to campaigns, we planted fake campaigns, built from the same hour's real answer times, to see whether the detectors could find them among real users.
\end{plainbox}

\begin{table*}[t]\centering\small
\caption{Locked predictions on the Chutes trace. W0 is the development window; W1 and W2 are held out, with code frozen. R2b, R6b and R9 were registered after W0.}\label{tab:chutes}
\setlength{\tabcolsep}{8pt}
\begin{adjustbox}{max width=\textwidth}
\begin{tabular}{@{}p{8.2cm}ccc@{}}
\toprule
Prediction & W0 (dev) & W1 & W2\\
\midrule
R1 random pools \texttt{none}: light $\ge95\%$ & 200/200 & 200/200 & 199/200\\
\quad heavy $\ge90\%$ & 194/200 & 199/200 & 192/200\\
R2 ring / successor $z>6$ in $\le1\%$ of pools & no (2.5\%) & yes (0\%) & \textbf{no} (4\%)\\
R2b same, pools with $\le3$ members of a real group & -- & yes$^\S$ & \textbf{no}$^\S$ (4\%)\\
R4 injected pools classified, $\ge10/11$ & 10/11 & 10/11 & 10/11\\
R5 spectral v0.8 finds injected ring among all & no & \textbf{no} & \textbf{no}\\
R6 discoverability, $\kappa=1.9$, $\ge75\%$ of cells & 4/6 & \textbf{2/4} & \textbf{4/6}\\
R6b same, $\kappa=3.2$ refitted on W0 & -- & 4/4$^\dagger$ & 6/6$^\dagger$\\
R7 expansion AUC $\ge0.85$ at 20\,min & .95--1.00 & .95--1.00 & .94--1.00\\
R9 real rotation groups have uniform counts & yes & n/a$^\ddagger$ & n/a$^\ddagger$\\
\bottomrule
\multicolumn{4}{@{}p{14.5cm}@{}}{$^\dagger$No cell recovered the ring within the hour, and all but one predicted that ($T^*>60$\,min); the exception ($T^*=56$\,min) is consistent under the factor-of-2 rule. $^\ddagger$Frozen site-wide discovery found no real group. $^\S$With no group found, R2b reduces to R2.}
\end{tabular}
\end{adjustbox}
\end{table*}

\textbf{What held.} False alarms stayed low (R1; Table~\ref{tab:chutes}). With its occupancy null re-estimated on each window's own random pools, frozen v0.7 labelled 199--200 of 200 random light pools and 192--199 of 200 heavy pools \texttt{none}. Classification held too (R4): ten of the eleven injected pools received the class they had received in the Two-replica run, in every window. Shared counters were classified as shared rings at ring $z=42$--46, per-worker counters as per-worker rotations at successor $z=59$--75, and pinned keys and the steady open loop as \texttt{none}. The one miss was the same pool in every window: a key held for 60\,s, which is short compared with an hour, was read as a shuffle. The result that matters most for this paper is that forensic expansion carried over unchanged (R7). From a 20\% seed, the shuffle, company, hashing and hedged pools ranked at AUC 0.94--1.00 within 20 minutes among 1{,}611--3{,}069 real accounts, in every window.

\textbf{What failed.} Blind discovery of the injected shared ring (R5) failed in all three windows, and so did the discoverability law with its simulation constant $\kappa=1.9$ (R6). Against 500 real accounts the law predicted recovery within 20--24 minutes. Instead, v0.8 recovered one of the two rings on W0 only at the full hour, and neither ring on W1 or W2. Real traffic is harder than our synthetic humans. On W0 the one recovered cell implies $\kappa\approx3.2$, which means windows about 2.8 times longer. With that refitted value (R6b), every held-out cell was consistent. This is a weak test, though, because all but one cell predicted no recovery within the hour, and none occurred. The real rings described below give a sharper check.

Surrogate calibration (R2) failed on heavy pools in W0 and W2. In W0 the excess came from random pools that happened to contain several members of a real 46-account rotation group, and no pool without a member fired. In W2 the firing rate rose with the number of members of a real 56-account group that a random pool contained. None of the 12 pools without a member fired, against 3 of the 10 pools with six or more. But 4 of the 97 pools with only one or two members also fired, so R2b failed whichever way it is read. Frozen discovery found no group in W2, which reduces R2b to R2 (4\%), and counting the post-hoc 56-account group, 4 of 151 pools with at most three members fired (2.6\%). The 99th percentile of ring $z$ in W2's heavy pools was 6.7, against 2.8--3.4 in both strata of W1 and in W2's light stratum. A threshold of 6 is therefore not portable from one window of heavy traffic to another. As WildChat had already suggested, a site should set it from its own past windows.

\begin{table*}[t]\centering\small
\caption{Groups of real accounts that share dispatch. ``Site'': found by frozen v0.8 over all accounts. ``Heavy'': found post hoc by v0.8 over heavy accounts only. Requests are completed requests per account in the hour (10th--90th percentile); $p$ is the lower tail of the $\chi^2$ dispersion test against random dispatch. Class: frozen v0.7 label (relay: shared round-robin, open-loop relay).}\label{tab:realgroups}
\setlength{\tabcolsep}{5pt}
\begin{adjustbox}{max width=\textwidth}
\begin{tabular}{@{}lcccccl@{}}
\toprule
Window & Found & Accts & Requests & $p$ & Models & Class (statistic)\\
\midrule
W0 & site & 46 & 185--190 & $1\times10^{-30}$ & 1 (96\%) & relay (ring $z=53.5$)\\
W1 & heavy & 12 & 1{,}194--1{,}205 & $4\times10^{-6}$ & 10 & per-worker (succ. $z=7.2$)\\
W1 & heavy & 8 & 1{,}138--1{,}204 & 0.72 & 6 & bunching ($R=2.9$)\\
W2 & heavy & 56 & 168--173 & $7\times10^{-34}$ & 1 (98\%) & relay (ring $z=42.0$)\\
W2 & heavy & 16 & 273 (all) & $\approx0$ & 16 & bunching ($R=10.8$)\\
\bottomrule
\end{tabular}
\end{adjustbox}
\end{table*}

\textbf{Real shared dispatch (R8, R9).} Production traffic does contain rotation pools. Table~\ref{tab:realgroups} lists every group of real accounts that spectral v0.8 found. In W0 the frozen site-wide scan, run over all 3{,}014 accounts, found 46 accounts that each completed 183--191 requests in the hour, almost all to one model. If requests were dealt out to 46 accounts at random, their counts would spread with $\chi^2\approx45$. The observed value is 0.89 ($p=1.4\times10^{-30}$). Counting every request started, including the few that never completed, the range is $189$--$192$ and $\chi^2=0.10$. A counter gives each account $n/M\pm1$ requests, and that is what we see. The frozen classifier, which had never been tuned on real data, labels the group a shared round-robin relay at ring $z=53.5$, and the group's top eigenvalues form the degenerate pair $(6.80, 6.34)$ that Proposition~\ref{prop:ring} predicts.

In W2 the frozen site-wide scan found nothing. Rerun post hoc over the 867 heavy accounts only, v0.8 found a second relay of the same kind: 56 accounts, 167--175 completed requests each ($\chi^2=1.4$ on 55 degrees of freedom), one model, ring $z=42.0$. In W1 the same post-hoc scan found 12 accounts that completed 1{,}188--1{,}222 requests each across 10 models; the classifier called it a per-worker rotation. The scans also turned up 16 accounts in W2 with exactly 273 requests each, every one on a different model and busy at the same moments. That is what a benchmark harness looks like when it sends each prompt to many models at once, and the classifier correctly called it bunching rather than rotation. A looser group of 8 accounts in W1, on 6 models and without uniform counts, was also called bunching.

Every group labelled a rotation had request counts more uniform than random dispatch allows, as R9 requires, although frozen site-wide discovery found no group in W1 or W2 and so R9 could not be tested there. The relays match the round-robin multi-key mode of open-source API gateways such as new-api~\cite{newapi}. We cannot see who runs these groups, and on an open inference platform sharing accounts this way may be entirely legitimate.
\begin{plainbox}\plainlead
Two independent kinds of evidence agree. The timing pattern says ``these accounts are fed by one round-robin counter'', and the request counts say ``these accounts received near-equal shares, within a few requests'', which is what a counter produces and chance almost never does.
\end{plainbox}

\textbf{The discoverability law on real rings.} The law also explains which real rings blind discovery found and which it missed. The constant $\kappa=3.2$ was fixed on W0's \emph{injected} cell before W2 was opened. With it, and with each relay's measured top eigenvalue ($\theta_1=6.80$ and 5.21) and the background's correlation time ($\tau_c=5.3$\,s), the law makes four predictions:
\begin{itemize}
\item W0 relay among the 891 heavy accounts: $T^*=18$ minutes; first recovered, whole and pure, at 20.
\item W0 relay among all 3{,}014 accounts: $T^*=59$ minutes; recovered at 60, with 20\% of it at 45.
\item W2 relay among the 867 heavy accounts: $T^*=29$ minutes; recovered at 60, just outside the factor of 2 (2.08).
\item W2 relay among all 3{,}069 accounts: $T^*=102$ minutes; not recovered within the hour, which is why the frozen scan missed it.
\end{itemize}

With the simulation constant $\kappa=1.9$, three of the four predictions would have been wrong, with the rings found 2.9--5.9 times later than predicted. The check is post hoc, but its constant was not fitted to these rings. The practical lesson is quantitative. On production traffic, the law says that blind discovery within an hour needs $\theta_1\gtrsim6.7$ among 3{,}000 accounts, or $\theta_1\gtrsim4$ among 1{,}000, and the W2 relay, at $\theta_1=5.2$ among 867 accounts, took twice its predicted time. Busy relays clear that bar; the lightly loaded closed-loop rings we injected ($\theta_1\approx2.7$) do not. Restricting the scan to one stratum, here the heavy accounts, shortens $T^*$ in proportion to $N$.

\textbf{Hand-off graph.} The v0.8 graph does not carry over to production traffic. Platform accounts end requests every few milliseconds, so they act as hubs and merge unrelated keys. The graph recovered the per-worker counter (100\%) and the company (90--100\%), but at most 25\% of every other injected pool, and it merged real accounts with injected ones. It was already the weaker detector in the Two-replica run (Section~\ref{sec:sitewide}). On real traffic it needs platform accounts removed, or a null that conditions on each key's own rate, before it can be used.

\subsection{An adaptive attacker on production background}\label{sec:adaptive}
The Evasion run measured the cost of evasion against synthetic users. Here we let an attacker who knows every detector choose its dispatcher, and measure how much throughput it keeps. Each configuration is simulated with W2's own answer durations, added to W2's 3{,}069 real accounts, and run through the whole frozen suite: the per-pool classifier with its occupancy, ring, successor and start-count statistics, failover after a refusal, and forensic expansion from a 20\% seed. A pool \emph{evades} if it is classified \texttt{none}, no statistic exceeds $z=3$ in its alarm direction, and expansion stays below AUC 0.7. Occupancy and start-count $z$ count in either sign. Ring, successor and failover $z$ count only as excesses, because shuffles and default retry-on-429 clients also produce negative values.

We give the attacker 40 keys and one of three budgets: 8 workers under a slack quota of 15 requests per minute per key; 8 workers under a binding quota of 3 per minute, where concurrency is scarce; and 40 workers under that same binding quota, where it is ample. The strategies are counters with up to 30\,s of jitter, keys held for 5--100 requests, shuffle, rotation on 429, pinned workers, pinned workers that pause between requests with gaps resampled from real light accounts, and Poisson open loops running at 50--95\% of the closed-loop rate with at most $K$ requests in flight. Pinned workers use one key each, so with 8 workers they hold only 8 keys.

\begin{table}[t]\centering\small
\caption{Adaptive attacker on the real W2 background. Each cell: throughput relative to the counter under the same budget, the frozen class, and the largest detector $z$ in its alarm direction (see text). Bold: evades every detector.}\label{tab:adaptive}
\setlength{\tabcolsep}{1.5pt}
\begin{adjustbox}{max width=\columnwidth}
\begin{tabular}{@{}lccc@{}}
\toprule
& $K{=}8$, 15/min & $K{=}8$, 3/min & $K{=}40$, 3/min\\
\midrule
counter & 1.00 ring (28) & 1.00 ring (40) & 1.00 relay (46)\\
counter, jitter 6\,s & 0.82 ring (34) & 0.80 ring (28) & 0.98 relay (36)\\
counter, jitter 30\,s & 0.49 ring (15) & 0.51 ring (11) & 0.65 relay (34)\\
hold 5 & 0.93 ring (6) & 1.00 ring (16) & 0.95 relay (26)\\
hold 100 & 0.87 sticky (4) & 0.71 sticky (3) & \textbf{0.79} none (1)\\
shuffle & 0.97 shuffle (6) & 0.89 shuffle (6) & 0.97 shuffle (3)\\
rotate on 429 & 0.93 p.-w. (13) & 0.99 ring (18) & 1.03 shuffle (4)\\
pinned & \textbf{0.98} none (1) & \textbf{0.61} none (2) & \textbf{0.83} none (1)\\
pinned, human gaps & \textbf{0.11} none (1) & \textbf{0.12} none (1) & \textbf{0.15} none (2)\\
open loop, 50\% & \textbf{0.50} none (0) & \textbf{0.50} none (1) & \textbf{0.45} none (1)\\
open loop, 80\% & 0.75 shuffle (3) & 0.76 shuffle (2) & \textbf{0.60} none (2)\\
open loop, 95\% & 0.89 shuffle (4) & 0.93 shuffle (3) & \textbf{0.66} none (0)\\
\bottomrule
\end{tabular}
\end{adjustbox}
\end{table}

Rotation is always caught (Table~\ref{tab:adaptive}). Every counter, at every jitter level up to 30\,s and in every budget, was classified as a shared ring or relay at ring $z\ge10$, so jitter only cost throughput. Escaping requires pinning, very long holds, or an open loop. Under a slack quota pinning is free, keeping 98\% of the counter's throughput, as Proposition~\ref{prop:imposs} implies. A binding quota with scarce concurrency is where evasion hurts most: the best evader keeps 61\% (pinned on 8 keys) and the best Poisson loop 50\%, which is Proposition~\ref{prop:pincer} measured on real noise. With ample concurrency the cost shrinks again. With 40 workers, pinning keeps 83\%, holding a key for 100 requests 79\%, and an open loop at 95\% of capacity 66\%. Pacing pinned workers like real users is ruinous, keeping only 11--15\%.

On W1's background (1{,}611 accounts) exactly the same configurations evaded in every budget, and no others did. Pinned workers kept 106\%, 55\% and 87\% of the counter's throughput in the three budgets, and every counter was again classified as a shared ring or relay (ring $z\ge8.5$). Across both windows, the best evasion under a binding quota with scarce concurrency keeps 55--61\% of throughput.
\begin{plainbox}\plainlead
An attacker can always hide by giving each worker its own key, but that has a price. When each key may only send a few requests a minute and the attacker cannot run many workers at once, hiding costs roughly 40\% of its speed. When either constraint is relaxed, hiding becomes much cheaper.
\end{plainbox}

One weak signal remains for pinned and held keys under a binding quota. After a refusal, other keys start \emph{less} often than expected ($z=-4.0$ to $-4.9$), because the worker retries its own key. Default SDKs do the same on a 429, so this does not separate a campaign from a benign client, and we do not count it. Expansion from a seed turned out to be the most fragile signal against an adaptive attacker. Two seconds of jitter cut the counter's AUC from 0.91--0.97 to 0.69, and a shuffle with as many workers as keys fell to 0.47. Its value lies against default tooling, which is what the documented campaigns used~\cite{anthropic2026}.

\section{Deployment: precision, analyst load and scale}\label{sec:deploy}
\textbf{Precision.} Across the 2{,}000 human key-runs of the Repeat, Stress, Evasion and Two-replica runs, spectral discovery, in every version, selected no human key in the reported runs. The 95\% upper bound on its per-key false-selection rate is therefore $3/2000\approx0.15\%$. Re-running the Two-replica analysis with 10 surrogate seeds per phase gave 2 human selections in 8{,}000 seed-by-human-key trials (0.025\%). The hand-off graph is weaker. Without rotate-on-429 pools (Repeat, Stress and Evasion runs) it selected no human. With them (Two-replica run), the frozen v0.4 graph selected 47 of 800 human key-runs and the v0.8 graph 6 of 800 (0.75\%), so a hand-off edge to a key that is held until refused should be confirmed, by the watermark or on a second window, before anyone acts on it.

Independent users are not where precision is lost. Legitimate shared dispatch is: a company spreading one queue over several keys (pool C) is detected exactly like a campaign. Suppose $c$ campaign pools and $b$ benign shared-dispatch pools across \emph{unlinked} accounts are active in a window. Blind discovery then has a precision of about $c/(c+b)$, which is 0.91, 0.5, 0.09 and 0.01 for $b/c=0.1$, 1, 10 and 100, and an analyst reviews $1+b/c$ groups per true campaign. Keys under one billing organisation are linked anyway, so only shared dispatch \emph{across} accounts (resellers, shared team credentials, aggregators) counts towards $b$. The Chutes trace gives a first measurement of it (Section~\ref{sec:chutes}). Each of the three production hours, among 1{,}611--3{,}069 active accounts, held one rotation group of real accounts (12--56 accounts; two found only post hoc). Two of the hours also held a group the classifier labelled bunching, one of them clearly a model-comparison fan-out. On an open inference platform, then, $b$ is about one rotation group per hour per few thousand active accounts. Whether those groups are benign we cannot tell, and at a provider whose accounts are bound to identities $b$ is probably lower. Either way, a blind alarm needs confirmation from a second channel before anyone acts on it; forensic expansion already starts from one.
\begin{plainbox}\plainlead
If honest shared setups outnumber campaigns ten to one, nine out of ten blind alarms point at honest users. That is why we recommend starting from an alarm that is already about extraction and using timing to find the rest, rather than using timing alone to accuse anyone.
\end{plainbox}

\textbf{Forensic expansion sidesteps the base rate.} Its entry condition is a precise alarm: content at 0.94--1.00~\cite{khilar2026}, or a payment or metadata link. Expansion then ranks keys given that alarm. With a 20\% seed, precision at pool size was 0.72 for the LiteLLM shuffle, 1.00 for the company, 0.59 for per-worker rotations and 0.29 for least-busy. Agents and pinned keys could not be expanded. A benign company is expanded as readily as a campaign, so expansion inherits the intent of its seed and should only start from an alarm that is itself about extraction.

\textbf{Timing.} Content classifiers ban accounts within about two hours~\cite{anthropic2026b}. Expansion reached AUC 0.86--0.98 within 10 minutes for shuffle and company pools in the Repeat run, and 0.85 for per-worker rotations within 20 minutes. In the frozen Two-replica test it reached 0.82--0.98 within 10 minutes and 0.92--1.00 within 20 minutes for shuffle, company, hashing and hedging. It therefore fits between the first alarm and the ban, and it finds accounts the classifier has not yet flagged. Blind discovery is the slow path. Proposition~\ref{prop:bbp} requires hours of data at $10^5$ keys, and on production noise its constant is about 1.7 times larger ($T^*$ 2.8 times longer; Section~\ref{sec:chutes}). It is meant for campaigns that content never flags and that persist, such as relays and covert prompts.

\textbf{Permutation budget.} Exact watermark p-values with $B$ permutations cannot go below $1/(B+1)$. The watermark is applied only to candidate pools, tens per day, and at $B=10^4$ each test takes seconds. Holm control over tens of tests needs a floor near $10^{-4}$, which $B=10^4$ provides. Site-wide permutation testing at $10^5$ keys is neither needed nor tractable: the passive scan uses its surrogate bulk edge, and the watermark confirms afterwards.

\textbf{What timing adds to linkage.} Payment and metadata links did connect real campaigns~\cite{anthropic2026}. Timing adds two things. It extends any such link, because a payment cluster is just another seed. And it reaches keys that were bought or stolen separately and share nothing with the seed except the client that drives them.

\section{Discussion}\label{sec:discussion}
\textbf{What an alarm means.} Timing proves that keys share dispatch machinery. It does not prove that they extract anything. We recommend using it for triage and forensic expansion behind a content alarm or human review, and not for automatic enforcement.

\textbf{Why ring order matters.} The recovered order is the client's own key list. It gives the worker count independently (Proposition~\ref{prop:succ}), shows which keys sit in a single configuration file, and lets an investigator predict which key a campaign will move to after a ban.

\textbf{What the attacker learns.} Every passive signature here is defeated by pinning or a steady open loop, and the watermark by 2\,s of jitter. So the value of timing lies against default tooling, which the documented campaigns used, and in the cost it imposes. An evading campaign gives up throughput (jitter), concurrency (Proposition~\ref{prop:pincer}) or the closed loop. On production background, with binding quotas and scarce concurrency, the best evasion kept 55--61\% of the throughput that rotation achieves (Section~\ref{sec:adaptive}).

\textbf{Limits of the testbed.} Our live results come from one 0.5B model on one GPU, synthetic clients, and runs of 40--60 minutes. The Stress run adds heavy-tailed lengths and a saturated KV cache, and the Two-replica run adds two replicas behind a provider-side balancer and per-key quotas. The Chutes windows add real users, real models and real replicas, but their campaigns are injected and each window lasts one hour. Production traffic is heavier and follows a daily cycle, and its replicas are many. Proposition~\ref{prop:bbp} implies hours of data at $10^5$ keys and about 34\,h at $10^6$, so at scale the binding constraint is statistical, not computational. Every campaign in our experiments is synthetic and was written by us, and the decision list was fixed on our own simulator. The Two-replica run's five unseen dispatchers are the only test of how it generalises. All five were flagged and both predicted classes held, but five is a small sample.

\textbf{Availability and disclosure.} The code for all ten experiments, the gateway, the client suite, the analysis library, the request logs of the live runs, and the locked predictions and frozen-code hashes of the Production study are available at \url{https://github.com/nssprogrammer/dispatcher_fingerprinting}. A file in the repository maps every table, figure and quoted number in this paper to the script that produces it. The evasions we measure, pinning and steady open loops, are already default behaviours of common tools. Publishing their cost tells providers which signals remain informative.

\textbf{Ethics.} All campaigns are synthetic. WildChat accounts are hashed IP addresses and Chutes accounts are anonymised, and we made no attempt to re-identify anyone. We do not claim that any real group we found is abusive.

\section{Conclusion}
The way a client hands its workers to API keys is written into the provider's logs, and real logs contain such pools: in three hours of a public serving trace, the frozen classifier named two round-robin relays. Shared counters leave a travelling wave whose shape and harmonics the theory predicts, per-worker counters leave fixed successors, and routers leave turn-taking on the timescale over which they hold a key. The most useful consequence is forensic. From a few accounts that a content alarm has already named, timing finds the rest of a router-driven campaign within minutes. Blind discovery covers persistent campaigns that content never sees, though on real traffic it reaches only strongly loaded rings within an hour, and a spread watermark makes linkage exact on suspects. The frozen detectors held under a saturated KV cache, on two replicas and under quotas, and on production traffic the classifier and forensic expansion held while blind discovery fell short. Only pinned keys and steady open loops escape, and timing provably cannot separate them from independent users. When quotas bind and concurrency is scarce, though, escaping costs the attacker roughly 40\% of its throughput.

\appendices
\section{Proofs and assumptions}\label{app:proofs}
Throughout, occupancy processes are taken to be stationary over the analysis window, and moments are time averages over it. By Bartlett's formula, the finite-window error of every sample variance and correlation used below is of order $\sqrt{\tau_c/T}$, where $\tau_c$ is the integrated squared autocorrelation time of high-pass occupancy (1.56\,s in the Repeat run).

\textbf{Turn-taking ratio.} Let $K$ always-active closed-loop workers each have a request in flight a fraction $p$ of the time, independently of each other, and let every request go to a uniformly random key. At each instant $N\sim\mathrm{Bin}(K,p)$, and $X_i\mid N\sim\mathrm{Bin}(N,1/M)$. Hence $\Var X_i=\E N\,\tfrac1M(1-\tfrac1M)+\Var N/M^2$, and summing over $i$ gives
$R=\Var N/\big(\E N(1-\tfrac1M)+\Var N/M\big)=(1-p)/(1-p/M)$. For open-loop Poisson arrivals, $N$ is Poisson (an $M/G/\infty$ system), and independent thinning makes the $X_i$ independent Poisson variables, so $\sum_i\Var X_i=\E N=\Var N$ and $R=1$. The intuition is that the total $N$ is limited by the number of workers, while each key's share of it still fluctuates; the more of the time the workers are busy, the less room the total has to move.

\textbf{Proposition~\ref{prop:ring}.} Assume $K<M$, so that a key never holds two requests, a fixed ring period $T_c$, and durations independent of everything else. Then $X_iX_{i+d}=1$ exactly when the two keys' on-intervals overlap, so $\E X_iX_{i+d}=\E O(S_d;D,D')/T_c$. Subtracting $\E X_i\E X_{i+d}=p^2$ and dividing by $p(1-p)$ gives $c(d)$. Because the dispatch process is stationary, $c$ depends on $d$ only, so the pool's correlation matrix is circulant up to noise of order $\sqrt{\tau_c/T}$. A circulant matrix is diagonalised by the Fourier vectors, and its eigenvalues are the discrete Fourier transform of its first row.

\textbf{Corollary (harmonics).} The profile $c(d)$ has an effective width of about $w_{\rm eff}=\lambda\E D$ keys, the number of dispatches during one answer. Its transform $\theta_k$ therefore decays over $k\approx M/(2\pi w_{\rm eff})$, and the narrower the profile, the more harmonics clear the bulk edge. In the Repeat run the shared counter has $w_{\rm eff}=9.5$ keys and $\theta_{1..3}=(7.09,1.81,0.22)$, the round-robin relay $w_{\rm eff}=4.6$ and $(4.86,2.30,0.85)$, and the jittered counter $w_{\rm eff}=3.1$ and $(2.88,2.34,1.39)$. Adding the BBP shift places the second harmonics of the jittered ring and the relay exactly at the Repeat-run outliers we could not explain at the time (3.35/3.22 and 3.01/2.94). This is why v0.4, which assigned keys to eigen-pairs, split the jittered ring in two. Version~0.5 selects keys on all outlier pairs and then splits them by correlation, which amounts to projecting on the multi-harmonic subspace of each ring.

\textbf{Proposition~\ref{prop:succ}.} Take a shared counter and a worker that dispatches at counter value $n$ and next at $n'$. In steady state, every other worker dispatches on average once between the worker's two dispatches, because all workers share the same cycle distribution. So $\E(n'-n)=K$. With per-worker counters, $n'=n+1$ by construction.

\textbf{Proposition~\ref{prop:K}.} This is the interactive response-time law $N=X(R+Z)$~\cite{denning1978}, with $N=K$ always-busy users, throughput $X=\lambda$, response time $R=\E D$ and think time $Z=\E G$. The excess of end$\to$start lags over a uniform-rate expectation follows the distribution of the worker's own gap, because other workers' starts are, to first order, uniform relative to any given end. Its mean estimates $\E G$.

\textbf{Worker-count intervals.} We split each run into six equal batches, estimate $K$ in each, and form a $t$ interval around the full-window estimate. With the truth in brackets, the Repeat run gave: shuffle 6.04 [5.87, 6.21] (6); least-busy 9.86 [9.80, 9.92] (10); shared counter 12.08 [12.05, 12.11] (12); per-worker 8.05 [7.86, 8.24] (8); counter + jitter 9.11 [7.83, 10.39] (8); company 3.96 [3.86, 4.07] (4); pinned 8.16 [8.02, 8.29] (8). The Stress run gave: shuffle 6.33 [6.12, 6.54]; least-busy 10.40 [9.60, 11.19]; shared counter 12.15 [11.98, 12.31]; per-worker 8.19 [8.09, 8.30]; counter + jitter 8.30 [8.01, 8.59]; company 4.36 [4.12, 4.61]; pinned 8.19 [7.96, 8.41]. The intervals cover the truth in 7 of 14 pool-runs. The misses are off by $-0.14$ to $+0.36$ workers, mostly upwards, which is more than the sampling error. The bias comes from reading $\E G$ at lag-bin midpoints and from queueing that the server adds to the gap. Agents (6 workers) gave 5.2 and 6.2. They are not always busy, because their tool calls count as gaps only when short. A locked prediction should therefore read $|\hat K-K|\le0.1K$ plus the interval, which every always-busy pool met.

\textbf{Proposition~\ref{prop:bbp}.} For a Wigner matrix with entry variance $\sigma^2$ plus a rank-$r$ spike of strength $\theta$, an outlier separates from the bulk if and only if $\theta>\sigma\sqrt N$, and it then sits at $\theta+\sigma^2N/\theta$~\cite{benaych2011}. Our matrix is a hollowed sample correlation matrix of dependent time series, so we use the same threshold with an empirical factor $\kappa$ (1.9 in simulation, about 3.2 on production traffic) that absorbs heterogeneous activity. The statement is therefore a calibrated approximation, not a theorem. The predicted outlier locations fell about 0.4 below the observed ones.

\textbf{Proposition~\ref{prop:wm}.} (i) The coins are drawn independently of all traffic, and $c_r$ is recorded before any delay. Under $H_0$ the joint law of $(c,s,a)$ does not depend on the coins, so given the data every relabelling of the coins with the same number of heads is equally likely. The rank of the observed $\hat T$ among the relabellings is then uniform, and the p-value is exact~\cite{fisher1935}. (ii) Within $[c_r,c_r+w)$, the worker's own next start lies at $c_r+b_r\delta+g$ with probability $q$, and background starts are uniform at rate $\lambda_{\rm bg}$. The expected share of in-window starts that come from the worker is $\pi=q/(q+\lambda_{\rm bg}w)$, and only that share moves, by $\delta$. (iii) The statistic is a difference of two means over about $n/2$ completions each, each mean with variance $\sigma_L^2/(n/2)$. Treating the in-window starts as uniform gives $\sigma_L\approx w/\sqrt{12(1+\lambda_{\rm bg}w)}$, which is within 5--23\% of the measured $\sigma_L$ for the four closed-loop pools of the Repeat run.

\textbf{Proposition~\ref{prop:imposs}.} In both cases the joint law of all keys' (start, end) processes equals that of the stated benign population. A level-$\alpha$ test rejects with probability at most $\alpha$ under that law, whatever statistic it uses. The cross-key watermark echo is zero in both cases, because no key's dispatch depends on another key's completions.

\textbf{Proposition~\ref{prop:pincer}.} (i) Each of the $n$ queries is carried by one key, so some key carries at least $n/M$. That key's statistic is about $\epsilon\sqrt{n/M}$, which stays below $z^*$ only if $M\ge n\epsilon^2/z^{*2}$. (ii) At most $K$ keys can carry a worker's first request. Every other key that carries any query receives its first query from a worker that started elsewhere, which is a switch, so if all $M$ keys are used there are at least $M-K$ switches. A closed-loop worker sends its next request after its previous answer ends, so a switch is an end on one key followed by a start on another, within the worker's gap. (iii) Combine (i) and (ii). The bound is tight for sequential retirement, in which each worker uses its keys one after another and never returns. The per-key monitor model $z\approx\epsilon\sqrt v$ holds for a mean shift per query, as in stateful monitors such as PRADA~\cite{juuti2019}. Monitors that need longer trajectories make the bound stronger.

\textbf{Cost corollary.} A Poisson count with mean $\mu$ per minute exceeds a cap $Q$ with probability about $\bar\Phi((Q-\mu)/\sqrt\mu)$. Keeping that probability at $\bar\Phi(z)$ requires $\mu\approx Q-z\sqrt Q$. In words, a sender who cannot see its own randomness coming must leave a safety margin of a few standard deviations below the cap.

\section{Each statistic as a score test}\label{app:score}
Approximate the high-pass occupancy vector of a pool by a Gaussian process with correlation $I+\epsilon A$, where $A$ describes the dispatcher. The score test for $\epsilon=0$ uses $\mathrm{tr}(A\hat C)$, with $\hat C$ the sample correlation. The four kinds of dispatcher give four tests. For \emph{random dispatch}, $A=\mathbf{1}\mathbf{1}^\top-I$ (every pair of keys coupled alike), and the score is the sum of off-diagonal correlations, a correlation-weighted form of $R-1$. For a \emph{shared counter}, $A$ is circulant with profile $c$ and an unknown key order. The generalised score maximises $\mathrm{tr}(P^\top A P\hat C)$ over permutations $P$, which is a quadratic assignment problem, and its spectral relaxation is the top eigen-pair of $\hat C$, that is, the ring statistic. For \emph{per-worker counters}, the alternative is a point-process cross-intensity concentrated on a permutation; its score is a maximum-weight matching on the excess hand-off matrix, and successor concentration is its row-wise relaxation. For \emph{pinned keys and open-loop random dispatch}, the alternative coincides with the null, so the Kullback--Leibler divergence between them is exactly zero (Proposition~\ref{prop:imposs}).

The statistics are therefore not ad hoc. Each is the locally most powerful test against the structure its dispatcher creates, with unknown orderings handled by relaxation. An exact likelihood ratio over the hidden assignment of workers to keys would be optimal only under its own model, and we leave it open.

\section{Matrix-free site-wide spectrum}\label{app:lanczos}
Let $X\in\mathbb{R}^{N\times T}$ be the sparse occupancy matrix, $H$ the block high-pass operator on time vectors, $D$ the diagonal of high-pass standard deviations, and $g$ the unit common-mode time vector. The hollowed matrix is $C=\tfrac1T ZZ^\top-\mathrm{diag}$, with $Z=D^{-1}XH(I-gg^\top)$. A product $Cv$ needs one sparse product with $X^\top$, one block-mean pass, one rank-one projection, and one sparse product with $X$. That costs $O(\mathrm{nnz}(X)+N+T)$ time and memory, and finding $k$ outlier eigenvalues by Lanczos iteration costs $O(k\cdot\mathrm{nnz}(X))$, without ever forming an $N\times N$ matrix. In the Repeat run ($N=703$, $T=21{,}800$, nnz $1.67\times10^6$) this reproduced the dense top-10 eigenvalues to three decimals in 0.7\,s. For the Chutes windows we built the hollowed matrix in chunks of keys, which matches the one-shot computation to within $10^{-15}$, and computed its top eigenpairs by Lanczos iteration whenever $N>1{,}500$; on our earlier runs this selected exactly the same keys and groups as a full eigendecomposition. At $N=10^6$ the binding constraint is statistical rather than computational: by Proposition~\ref{prop:bbp}, a strong ring among $10^6$ keys needs about 34\,h of data.

\section{Post-hoc analyses (not frozen)}\label{app:posthoc}
\begin{table*}[t]\centering\small
\caption{Every locked prediction and its outcome.}\label{tab:ledger}
\begin{adjustbox}{max width=\textwidth}
\begin{tabular}{@{}lp{11.5cm}c@{}}
\toprule
Run & Prediction & Held\\
\midrule
WildChat & 6-h pools: $R_{60}<0.85$ in $\le$1\%: random / country / browser+country / heaviest & 3 of 4\\
Pilot & occupancy $R_{20}<0.9$ for the script and the LiteLLM pool (go/no-go) & 1 of 1\\
Baseline & classification $\ge9/11$; ring order; worker count $\pm25\%$; controls at null & 4 of 4\\
Baseline & site graph recovers the round-robin relay & no (15\%)\\
Repeat & pinned: same-key echo only; humans calibrated; halves linked; classification $\ge10/12$; sideband graph & 5 of 5\\
Repeat & least-busy echoes on other keys & no (same key)\\
Repeat & random relay does not echo & no ($p=0.0015$)\\
Repeat & spectral v0.4 recovers the jittered ring & no (75\%)\\
Stress & stress real; classification v0.5, v0.6; v0.6 spectral; relay null; nulls calibrated; worker count & 7 of 7\\
Stress & slow rings lost at the 20\,s scale (revised after its rehearsal) & no\\
Stress & v0.5 spectral recovers all three rings & no (jitter 0\%)\\
Stress & sideband graph recovers the shared ring & no (35\%)\\
Evasion & ring under jitter; held keys; classification; spread watermark; visibility; bursty bunching & 6 of 6\\
Evasion & steady open loop classified none & no (bunching)\\
Two-replica, balanced & M1 replica null: $R$, ring, $\le2$ humans selected & yes\\
Two-replica, affinity & M1 replica null: $\le2$ humans selected & no (v0.8 graph 4)\\
Two-replica, affinity & M2 same-replica human pools at the null & yes\\
Two-replica, both phases & M3 references; M4 unseen dispatchers flagged, hash $\to$ shuffle, TTL $\to$ sticky; M5 failover after 429, none for pinned; M6 v0.8 ring, v0.7 fails; M7 pinned $\le0.25$/s; M8 expansion AUC $\ge0.85$; M9 v0.4 graph humans linked only to 429 pools & 14 of 14\\
Production, W1 and W2 & R1 random real pools \texttt{none}; R4 injected classification $\ge10/11$; R7 expansion AUC $\ge0.85$; R6b refitted law (one-sided) & 8 of 8\\
Production, W1 & R2, R2b surrogate calibration & 2 of 2\\
Production, W2 & R2, R2b ring $z>6$ in $\le1\%$ of heavy pools & no (4\%, 4\%)\\
Production, W1 and W2 & R5 injected ring found site-wide & no (0\%)\\
Production, W1 and W2 & R6 discoverability law with $\kappa=1.9$ & no (2/4, 4/6)\\
\midrule
Total & & 52 of 68\\
\bottomrule
\end{tabular}
\end{adjustbox}
\end{table*}

\textbf{Bulk-edge inflation and trimming (Evasion run, Two-replica rehearsal).} At the 20, 60 and 180\,s scales, the surrogate bulk edge in the Evasion run was 4.6, 5.9 and 8.3, against 2.8 in the Repeat run. The keys with the longest autocorrelation were those held for 5 requests (E5), where a worker keeps one key busy for about 13\,s. Such keys inflate the variance of null cross-correlations, and with it the null edge. Dropping keys whose integrated squared autocorrelation time exceeded the site median by more than 5 robust standard deviations (v0.7) removed the 20 E5 keys and 2 humans. The edge fell to 2.6, and the scan recovered all five shared-counter rings (jitter 0, 2, 6 and 15\,s, and 24 workers) at 20/20, pure.

v0.7 then failed its first test, on the simulated rehearsal that precedes every live run. In the Two-replica run the shared-counter ring has 40 keys visited every $\approx20$\,s, so each key's autocorrelation has a peak every period. The squared sum over 120\,s counted all of those peaks. The ring's own keys therefore had the longest correlation times (median 9.7\,s, against 3.3\,s for humans), and v0.7 trimmed 20 of them. Integrating the autocorrelation only up to its first zero crossing (v0.8) measures positive memory, which held keys have and ring keys do not. In the Evasion run, v0.8 trims exactly the 20 E5 keys and 2 humans and keeps all five rings at 20/20. On the 10-minute rehearsal it trims the rotate-on-429 keys (which are held until refused) and no ring keys, and recovers the ring at 40/40 in one phase and at 18/40, split in two, in the other. On a full 40-minute simulated phase, v0.8 recovered the ring at 40/40, pure, with no humans, while v0.7 trimmed 103 keys and recovered none of it. We fixed v0.8 before the live Two-replica run and report v0.7 there unchanged.

\textbf{Hand-off graph (Two-replica rehearsal).} On the same 40-minute phase the frozen v0.4 graph selected 10 humans, every one linked to a rotate-on-429 key. At the pool level, human starts did not follow those pools' ends any more often than random times did (8.4\% against 8.7\% within 50\,ms), so the links were false. Two things produced them. A rotate-on-429 key carries bursts of about 15 requests before it is refused, and the sideband null counts one human start after such a burst as several hand-offs. And the retest is Bonferroni-corrected over the screened edges only, although the screen used the same counts. The v0.8 graph counts distinct starts against the locally covered share of the sideband, which on its own left 4 humans, and corrects over all $N(N-1)/2$ pairs, which left none. Screening on one half of the window and testing on the other is also valid, but lost more recall. The correction costs recall on weak edges. Across the Repeat, Stress and Evasion runs and the rehearsal, v0.8 selected 0 humans, and recovered per-worker rotations (100\%), the rotate-on-429 pools (100\%), the clean shared ring (100\% in the Repeat run, 85\% in the rehearsal) and held keys (E5 100\%). It lost the jittered ring (78\% to 5\%) and most of the company in the Repeat run (70\% to 20\%), which spectral discovery and expansion cover.

\textbf{Hand-off graph on the live Two-replica run.} We re-ran the graphs from the released request logs and reproduced the counts from the original analysis: v0.8 linked 2 and 4 humans, and v0.4 linked 25 and 23 (24 and 23 in the original analysis). Every v0.8 human edge ran from the human to a rotate-on-429 key, with 8--14 starts on that key following the human's 5--7 ends within 1\,s, against 0.2--0.9 expected. The mechanism is the v0.4 one, now on the start side. When a key is refused, all workers move to the next key within a second, so its starts arrive in stints, and one chance coincidence between a human's end and the onset of a stint counts once per start. Collapsing each key's starts to bursts before the coverage test (v0.9, post hoc) removed every human edge in both phases, in the rehearsal, and in the Repeat, Stress and Evasion runs. It also removed most of the rotate-on-429 pools (recall 100\% to 2--5\%), while keeping per-worker rotations, the clean ring and held keys. A graph triggered by the provider's own refusals (key $i$ refused, then key $j$ starts within 0.3\,s) recovered the per-worker rotate-on-429 pool (100\%, purity 0.93) but only 5\% of the shared-pointer pool, and still linked 2 humans. Our conclusion is that, pair by pair, the evidence linking two keys of a rotate-on-429 pool consists of a few stint changes, and a chance coincidence with the start of a stint looks just the same. Those pools are caught reliably one pool at a time, by turn-taking ($z_R\approx22$), successor concentration and failover ($z=20$--55). To be found among all keys they need a seed or the watermark, not the pairwise graph.

\textbf{Start-count ratio.} Computing the same variance ratio on request \emph{start} counts, in 2\,s bins with 60\,s high-pass blocks, never uses durations, so batching cross-talk cannot affect it. Against 100 human pools, the steady open loop and the random relay returned to $0.98$ and $1.01$ ($z=0.4$, $-0.5$), and the bursty loop stayed bunched ($3.77$, $z=-68$). Closed loops were still detected, at lower power: $z=2.2$--$4.0$ in the Stress run and $4.0$--$10.7$ in the Evasion run. Least-busy and agents were not detected. We recommend it as a confirmation step for bunching alarms, not as a replacement for the occupancy ratio.

\textbf{Seed expansion (Repeat run).} The linking score is chosen per seed without labels: occupancy anti-correlation with the seed, symmetric hand-off excess, or positive co-movement, whichever ranks held-out seed members highest. Reported AUCs average 20 random seeds per setting.

\section{Locked-prediction ledger}\label{app:ledger}

The Production study counts only the held-out windows W1 and W2. W0 was the development window, and its outcomes are reported in Table~\ref{tab:chutes} but not counted. R3 (replica coupling) was declared not testable before W1 and W2 were opened. R9 could not be tested on W1 or W2, because frozen site-wide discovery found no real group there. The lock, both amendments, the data hashes and the frozen-code hashes are released with the code.

\end{document}